\documentclass[letterpaper, 10 pt, conference]{ieeeconf}  %

\IEEEoverridecommandlockouts                              %

\usepackage{graphicx}
\usepackage{xcolor}
\usepackage{amsmath}
\usepackage{hyperref}
\usepackage{amssymb}
\usepackage{booktabs}  %
\usepackage{multicol}  %
\usepackage{multirow}

\title{\LARGE \bf
RGB-Only Reconstruction of Tabletop Scenes \\
for Collision-Free Manipulator Control
}

\author{Zhenggang Tang,$^{1,2}$ Balakumar Sundaralingam,$^{1}$ Jonathan Tremblay,$^{1}$ Bowen Wen,$^{1}$ Ye Yuan$^{1}$ \\
Stephen Tyree,$^{1}$ Charles Loop,$^{1}$ Alexander Schwing,$^{2}$ Stan Birchfield$^{1}$
\\ $^{1}$NVIDIA%
\\ $^{2}$University of Illinois Urbana-Champaign%
\thanks{Work was completed while the first author was an intern at NVIDIA.}%
}

\begin{document}

\maketitle
\thispagestyle{empty}
\pagestyle{empty}

\newlength{\mycharwid}  %
\settowidth{\mycharwid}{8}

\begin{abstract}
We present a system for collision-free control of a robot manipulator that uses only RGB views of the world.
Perceptual input of a tabletop scene is provided by multiple images of an RGB camera (without depth) that is either handheld or mounted on the robot end effector.
A NeRF-like process is used to reconstruct the 3D geometry of the scene, from which the Euclidean full signed distance function (ESDF) is computed.
A model predictive control algorithm is then used to control the manipulator to reach a desired pose while avoiding obstacles in the ESDF.
We show results on a real dataset collected and annotated in our lab. Our results are also available at~\href{https://ngp-mpc.github.io/}{https://ngp-mpc.github.io/}.

\end{abstract}

\section{INTRODUCTION}

Robotic control systems often assume perfect knowledge of the scene for collision avoidance, planning, and control~\cite{schulman2014motion,morgan2021vision}. 
Such knowledge is difficult to acquire robustly in unstructured environments, because many long-standing problems in computer vision remain unsolved.
Moreover, the information bottleneck between perception and planning creates difficulty in using motion planners in unstructured real environments.

Previous attempts to build a  model of the scene often use  a voxel map representation~\cite{mainprice2016warping,muglikar2020voxel,mitash2020task}.
Voxel maps require the scene extent and resolution to be decided beforehand, and they are memory intensive, which limits their resolution. Although these issues can be alleviated to some extent via octrees~\cite{takikawa2021neural,yu2021plenoctrees} and/or hash functions~\cite{niessner2013real,mueller2022siggraph:instantngp}, such representations still suffer from fixed discretization. In practical applications, voxel-based methods have been sufficient to avoid obstacles. However, when it comes to navigating in tight spaces between obstacles or tasks where the robot has to move very close to an object (e.g., grasping), the limited resolution can prevent the task from being completed without collision.

Many systems require a depth camera for sensing.
But depth cameras exhibit noise that is difficult to model \cite{sterzentsenko2019self,chaudhary2016noise,sweeney2019icra,mallick2014sensors,haider2022sensors,handa:etal:2014,baruhov2020gan,nguyen2012kinect,iversen2017kinect}, and they struggle to handle transparent/reflective surfaces \cite{sajjan2020clear}, dark surfaces, and occlusions.
Due to these errors or occlusions, depth cameras often exhibit large holes in the output \cite{gu2017learning}, thus making accurate downstream processing more difficult.
With the increased progress of deep neural networks, there is reason to believe that RGB images can be used to estimate depth as well as, or perhaps even better than, off-the-shelf depth sensors.
After all, compared with triangulation-based depth sensors, multi-view RGB images provide mathematically identical information; moreover, RGB cameras are generally more ubiquitous, cheaper, and higher resolution, which provides opportunities for deep networks to memorize inductive depth biases. 
Indeed, in some cases, deep networks have been shown to estimate depth from multiple RGB images better than the output of depth cameras~\cite{lin2021iros:mvml}. 

\begin{figure*}
   \def\figoverviewscale{0.38}
    \centering
    \begin{tabular}{c}
    \includegraphics[scale=\figoverviewscale]{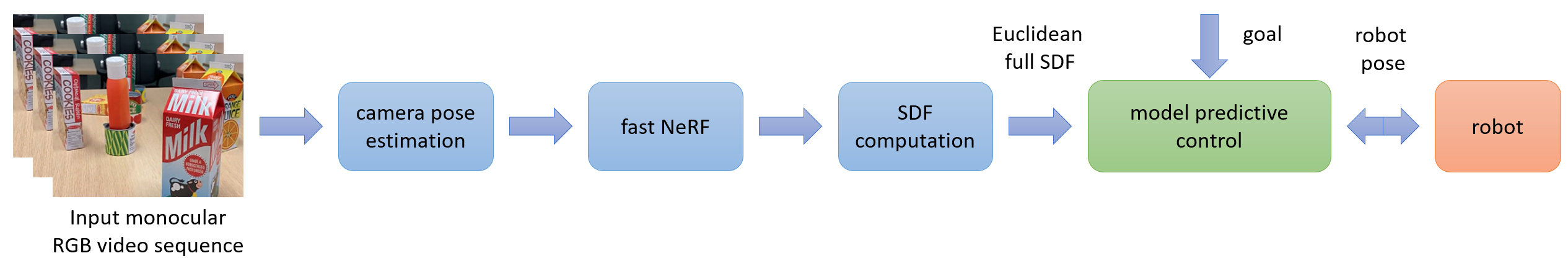}
    \end{tabular}
  \caption{System overview.  From an input monocular RGB video sequence, we infer a Euclidean full SDF by first estimating camera poses~\cite{schoenberger2016cvpr:sfm,schoenberger2016eccv:mvs}, running a fast version of neural radiance fields~\cite{mueller2022siggraph:instantngp}, marching cubes~\cite{lorensen1987siggraph:marchcube}, and SDF estimation.  The SDF is then fed, along with a goal, to the motion optimization which then controls the robot to reach the goal while avoiding obstacles. 
  }
  \label{fig:overview}
\end{figure*}

To explore this potential, we present a system for collision-free motion generation of a tabletop robot manipulator based on 3D reconstruction of a tabletop scene observed by an RGB camera from multiple views. 
Different from prior work, the primary novelty of our method is to build a \emph{Euclidean full signed distance function (ESDF)}~\cite{oleynikova2017iros:voxblox} representation of the geometry of the scene from only RGB images, which is then used for reactive control. 
Concretely, in this work we explore the possibility of extracting the ESDF representation of the scene from a learned neural radiance field (NeRF) encoding~\cite{mildenhall2020eccv:nerf}.

The method involves moving an RGB camera around the scene (either manually or via the robot hand) and recording images. 
The camera poses are registered for each image using existing simultaneous localization and mapping (SLAM) methods~\cite{schoenberger2016cvpr:sfm,schoenberger2016eccv:mvs}. 
The RGB images and camera poses are then fed to a fast version of NeRF~\cite{mueller2022siggraph:instantngp} to build a 3D model of the scene. 
From this model, a mesh is extracted, which is then used to compute an ESDF.
This ESDF is then plugged into a 
GPU-accelerated model predictive control method \cite{bhardwaj2021corl:storm} to generate reactive motions for the robot.

Compared to baseline methods, we show that our NeRF-based RGB-only reconstruction of the scene achieves significant improvement both in terms of geometric accuracy and task success.
To summarize, our contributions are as follows:
\begin{itemize}
    \item We propose a method to extract an accurate ESDF of a tabletop scene using a fast version of NeRF.
    \item Using this approach, we connect perception with control to facilitate the collision-free motion of a manipulator in an unstructured scene.
    \item We validate our approach on real data, showing considerable improvement over state-of-the-art methods.  We will share our annotated dataset with the community.
\end{itemize}

\section{RELATED WORK}

In the following we briefly review methods to compute classical and neural 3D scene representations, after which we describe their use in motion planning and control.

\textbf{Classic 3D scene reconstruction.} 
Various systems have been proposed to reconstruct the 3D geometry of a tabletop scene for manipulation.
CodeSLAM~\cite{bloesch2018cvpr:codeslam} focuses on reconstruction, but it does not use the reconstruction for control.
MoreFusion~\cite{wada2020cvpr:morefusion} fits known 3D models to the point cloud data.
Lin et al.~\cite{lin2021iros:mvml} describe a multi-level perception system. However, none of these methods connect the resulting reconstruction to motion planning. Their scene representations such as point clouds are also not well suited for motion planning which requires fast computation of collisions.

\textbf{Neural 3D scene representations.} 
With the growing success of neural radiance fields (NeRFs)~\cite{mildenhall2020eccv:nerf}, neural implicit representations of 3D scenes have gained attention for their ability to generate high-quality observations of the scene from novel views. NeRF-based  methods typically use a neural network to map 3D coordinates into values of density and radiance, which are used in differentiable rendering and allow end-to-end learning. Recent works have advanced neural 3D scene reconstruction in various ways:  Points2NeRF~\cite{zimny2022arx:pointstonerf} learns a NeRF representation from a colored point cloud.
Point-NeRF~\cite{xu2022cvpr:pointnerf} speeds up the training of NeRF by leveraging a pre-trained network, as well as pruning and growing of points. Instant-NGP~\cite{mueller2022siggraph:instantngp} drastically improves the speed of NeRF up to several orders of magnitude using a hash table. Besides novel view synthesis, Mescheder et al.~\cite{mescheder2019cvpr:occnet} also use a neural implicit representation to approximate the scene geometry with an occupancy network.

Alternatively, a signed distance field (SDF) is another  neural scene representation that is more suitable for reconstructing high-quality scene geometry. For instance,  Voxblox~\cite{oleynikova2017iros:voxblox} is a classic method for computing a Euclidean full SDF (ESDF) from quadrotor vision data. 
The authors of iSDF~\cite{ortiz2022rss:isdf} implement a framework to learn the full signed distance function (SDF) from depth images and camera poses, making note of the inadequacy of truncated SDFs for robotic manipulation. 
VolSDF~\cite{yariv2021neurips:volsdf} represents geometry, as the transformation of the SDF rather than the density. 
Similarly, NeuS~\cite{wang2021neurips:neus} represents the scene using a transform of the SDF.
Neural RGBD~\cite{azinovic2022cvpr:nrgbd} represents the scene using a projected TSDF from RGBD images.
NGLOD~\cite{takikawa2021cvpr:nglod} also represents the scene using an SDF, but focuses on fast rendering rather than reconstruction.
MonoSDF~\cite{yu2022arx:monosdf} exploits monocular cues in reconstructing the scene from multiple RGB views via the SDF.
MetaSDF~\cite{sitzmann2020neurips:metasdf} also uses NeRF to represent the scene as an SDF, utilizing meta-learning to speed up inference.
DeepSDF~\cite{park2019cvpr:deepsdf} learns a neural network to approximate the \emph{truncated SDF (TSDF)}, which, as we have already mentioned, unfortunately cannot be used for robot control outside the truncated region.
Different from prior work, our system performs 3D reconstruction a scene using multi-view RGB images by estimating the ESDF.

\textbf{Robot motion planning and control.}
A variety of methods have used 3D scene representations for motion planning and control. For example, 
Riemannian Motion Policies (RMPs)~\cite{ratliff2018arx:rmp} use a model-based approach in which a 3D model of the object of interest is known beforehand.
STORM~\cite{bhardwaj2021corl:storm} implements sampling-based model predictive control on a GPU and shows obstacle avoidance in a known world (primitive shapes).
Battaje et al.~\cite{battaje2022iros:ooaat} use a wrist-mounted RGB camera to focus on a single object in the world to avoid the obstacle. 
Yang et al.~\cite{yang2022icra:mpc} use model predictive control for reactive human-to-robot handovers.
Adamkiewicz et al.~\cite{adamkiewicz2022icra:vorn} use an existing NeRF model of the world to plan motions for a quadrotor. 
Driess et al.~\cite{driess2021corl:lmaf} describe an optimization-based manipulation planning approach via SDFs.
DiffCo~\cite{zhi2022trob:diffco} provides differentiable collision detection.
Danielczuk et al.~\cite{danielczuk2021icra:collfun} also describe neural collision detection.
Li et al.~\cite{li2021corl:visuomotor} learn a viewpoint-invariant 3D-aware scene representation via NeRF with time-contrastive learning for visuomotor control of a manipulator.
Along this line, Driess et al.~\cite{driess2022reinforcement} leverage a NeRF to learn a latent scene representation of the scene, which is then used to train an reinforcement learning (RL) policy for robot control. In closely related work, Pantic et al.~\cite{pantic2022icraw:sfog} use the internal representation of NeRF to approximate an ESDF, which is then integrated into an RMP framework for motion planning.

In contrast, our method enables robust and accurate ESDF reconstruction by first using a NeRF to reconstruct a detailed scene mesh. This allows our approach to be tested on real-world scenarios and be effective for manipulator control.

\section{METHOD}
Our approach takes as input a set of RGB views of the scene and builds a NeRF model from which a mesh is extracted, as described in Sec.~\ref{sec:rgbngp}. 
The ESDF is then computed from the reconstructed mesh and used with model predictive control to generate commands for the robot, as discussed in  Sec.~\ref{sec:mpc}.

\begin{figure}
   \def\figtabletopxscale{3cm}
    \centering
    \begin{tabular}{cc}
    \hspace{-2.1ex} \includegraphics[height=\figtabletopxscale]{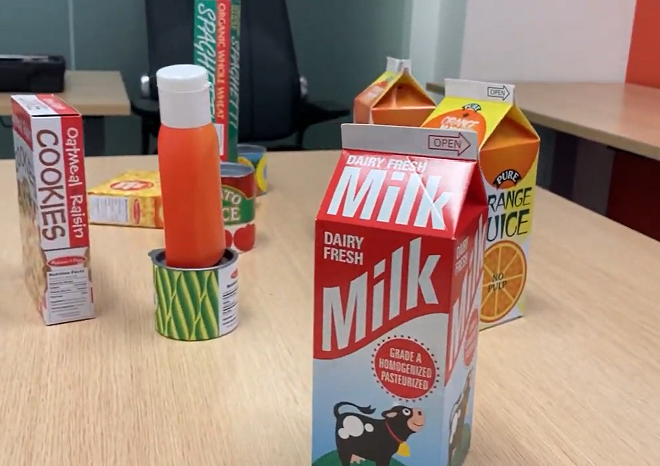} &
    \hspace{-2ex}
    \includegraphics[height=\figtabletopxscale]{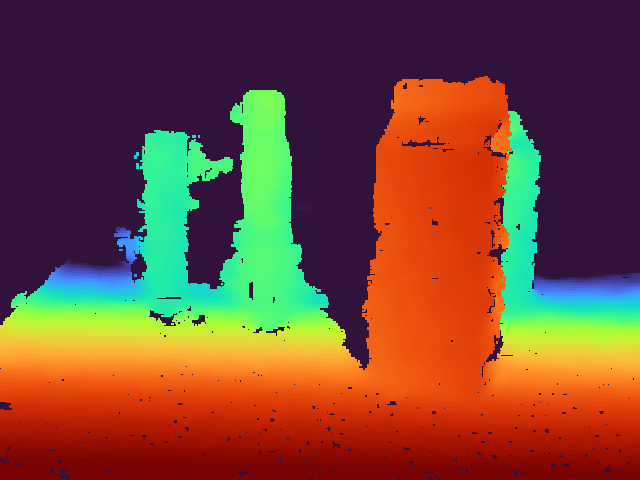} \\
    \hspace{-2.1ex}
    \includegraphics[height=\figtabletopxscale]{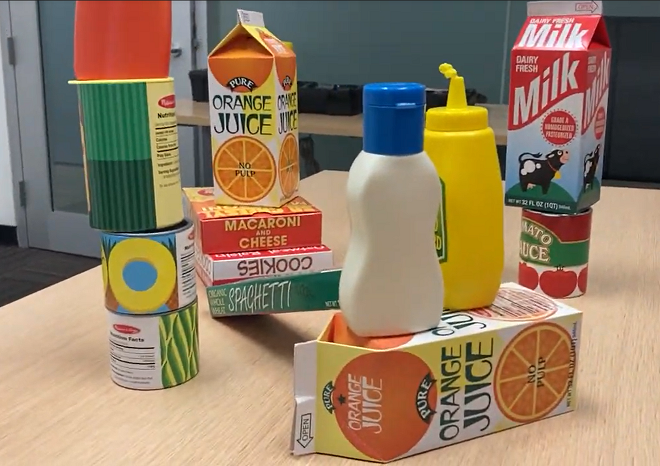} &
    \hspace{-2ex}
    \includegraphics[height=\figtabletopxscale]{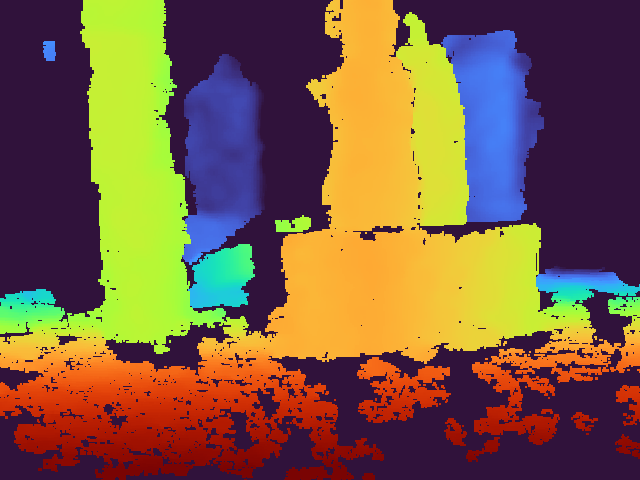} \\
    \hspace{-2.1ex}
    \includegraphics[height=\figtabletopxscale]{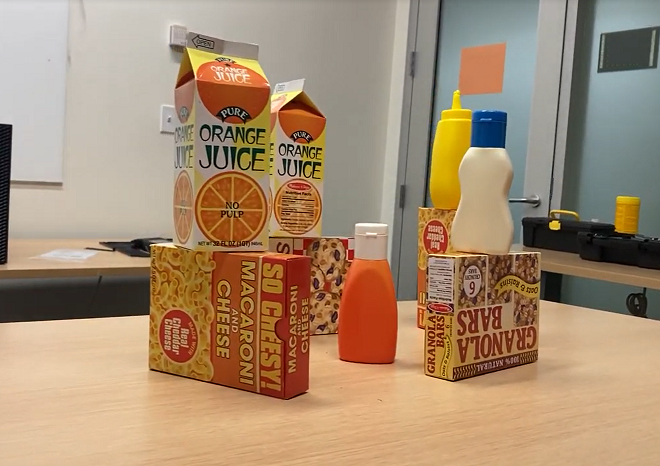} &
    \hspace{-2ex}
    \includegraphics[height=\figtabletopxscale]{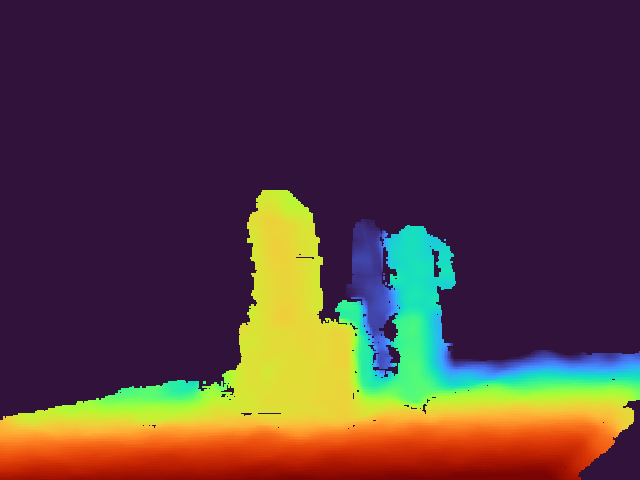}
    \end{tabular}
  \caption{Our dataset consists of images of three real-world tabletop scenes. {\sc Top to bottom:}  Scene 1, Scene 2, Scene 3.  {\sc Left:}  RGB images from the iPhone camera.  {\sc Right:}  Similar-viewpoint depth images from the RealSense camera (used only for computing ground truth and comparison, not used in our approach).  Depth maps are colorized with Turbo~\cite{mikhailov2019:turbo}.}
  \label{fig:tabletopx}
\end{figure}

\begin{figure*}
   \def\figzerolevelmeshscale{0.8in}
   \def\figzerolevelmeshspace{-2ex}
    \centering
   \begin{tabular}{ccccc}
    \includegraphics[scale=0.24]{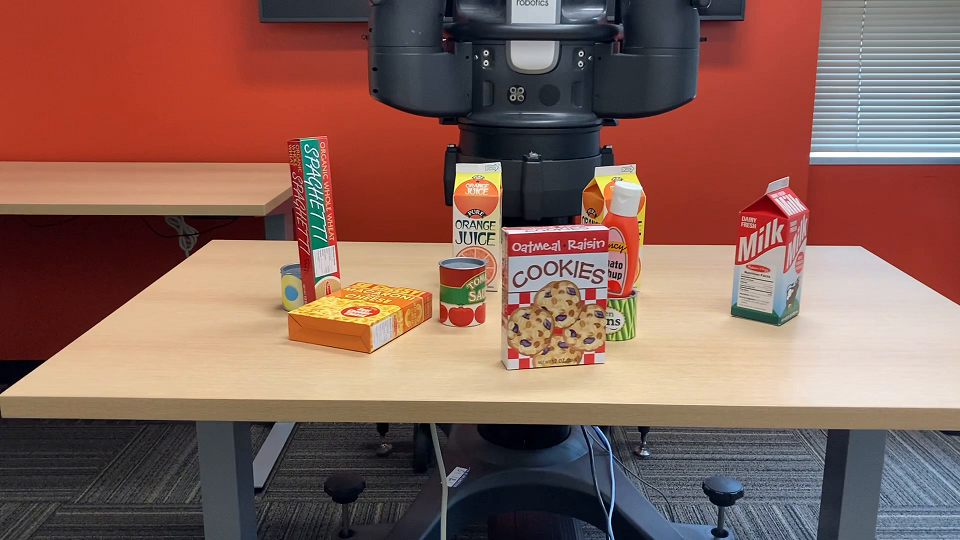} 
    &
    \hspace{\figzerolevelmeshspace}
    \includegraphics[height=\figzerolevelmeshscale]{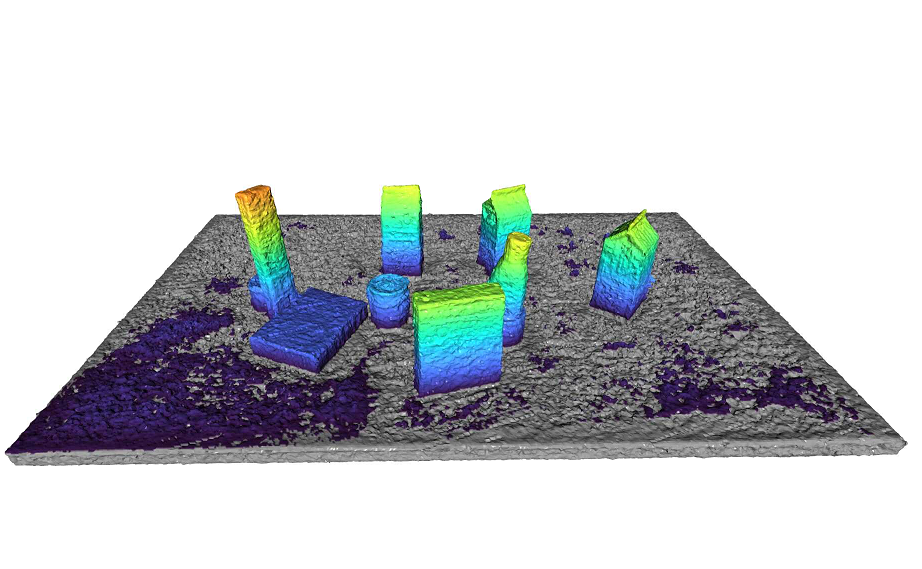} 
    \hspace{\figzerolevelmeshspace}
    &
    \hspace{\figzerolevelmeshspace}
    \includegraphics[height=\figzerolevelmeshscale]{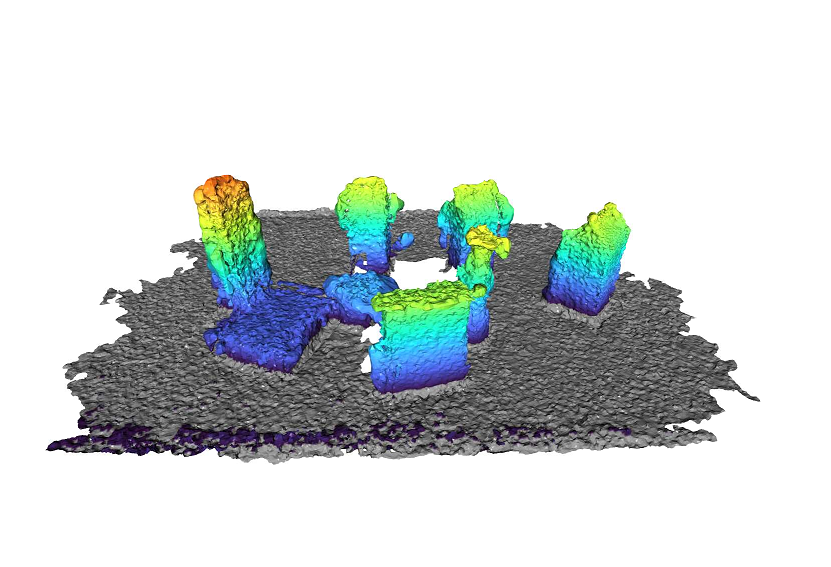} 
    \hspace{\figzerolevelmeshspace}
    &
    \hspace{-1.8ex}
    \includegraphics[height=\figzerolevelmeshscale]{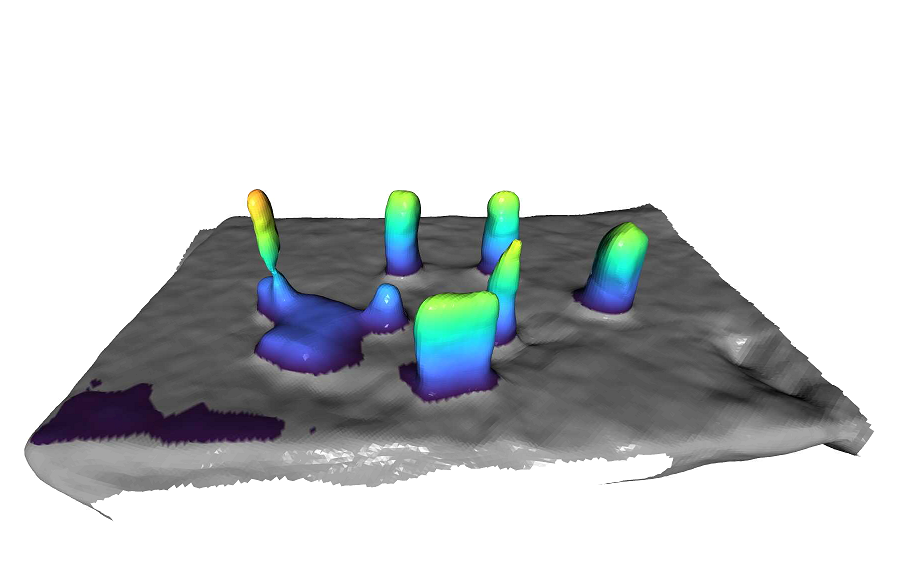} 
    \hspace{-1.9ex}
    &
    \hspace{\figzerolevelmeshspace}
    \includegraphics[height=\figzerolevelmeshscale]{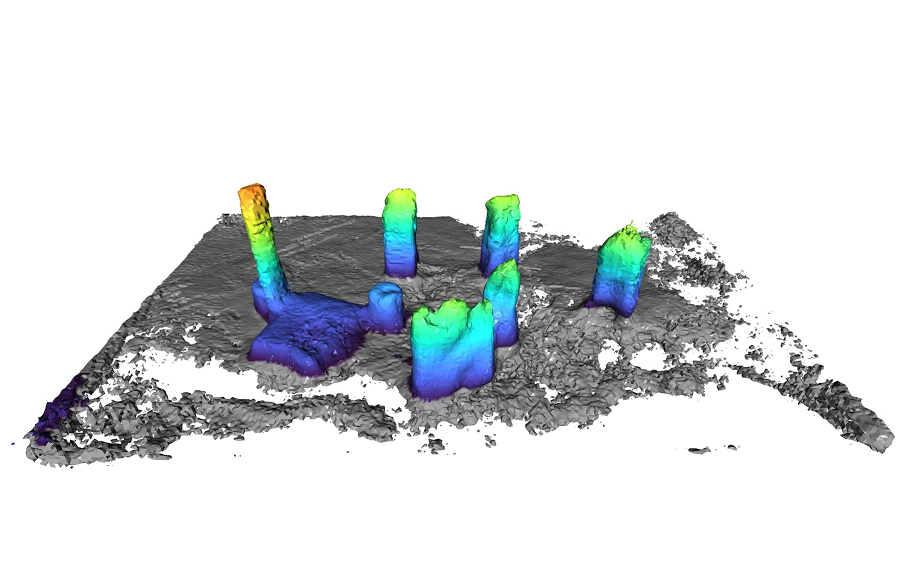}
    \hspace{\figzerolevelmeshspace}
\\
    \includegraphics[scale=0.24]{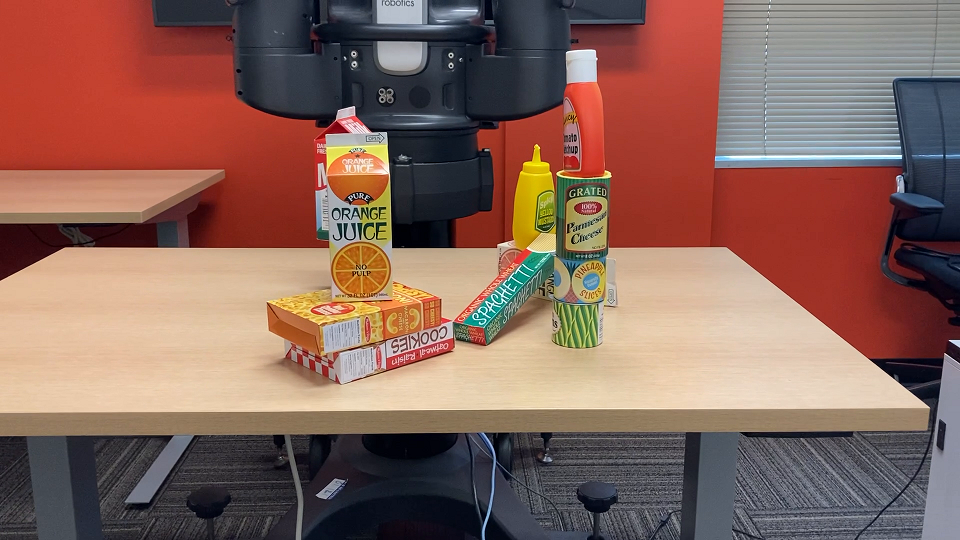} 
    &
    \hspace{\figzerolevelmeshspace}
    \includegraphics[height=\figzerolevelmeshscale]{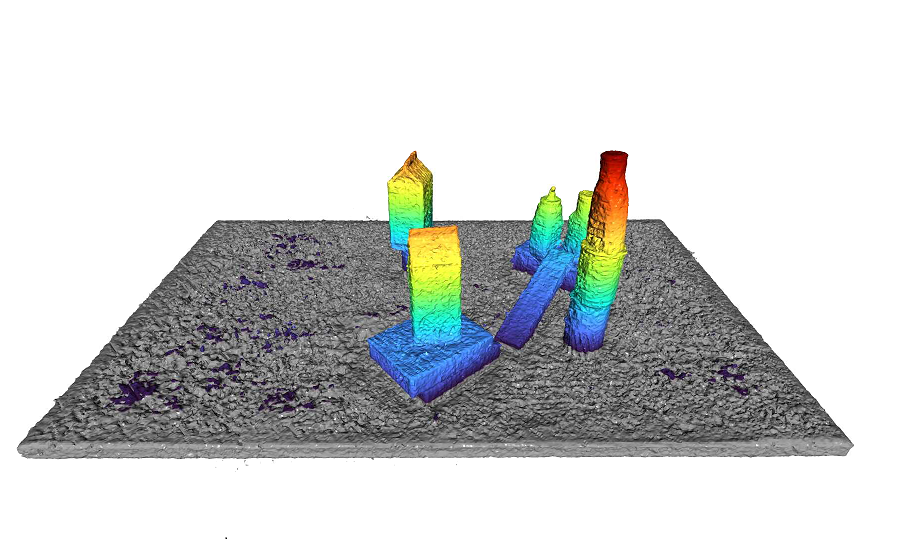} 
    \hspace{\figzerolevelmeshspace}
    &
    \hspace{\figzerolevelmeshspace}
    \includegraphics[height=\figzerolevelmeshscale]{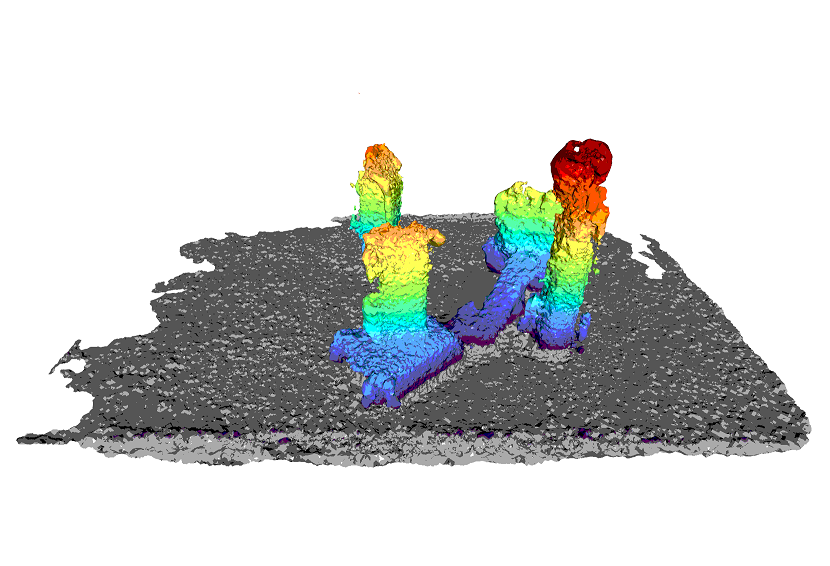} 
    \hspace{\figzerolevelmeshspace}
    &
    \hspace{-1.8ex}
    \includegraphics[height=\figzerolevelmeshscale]{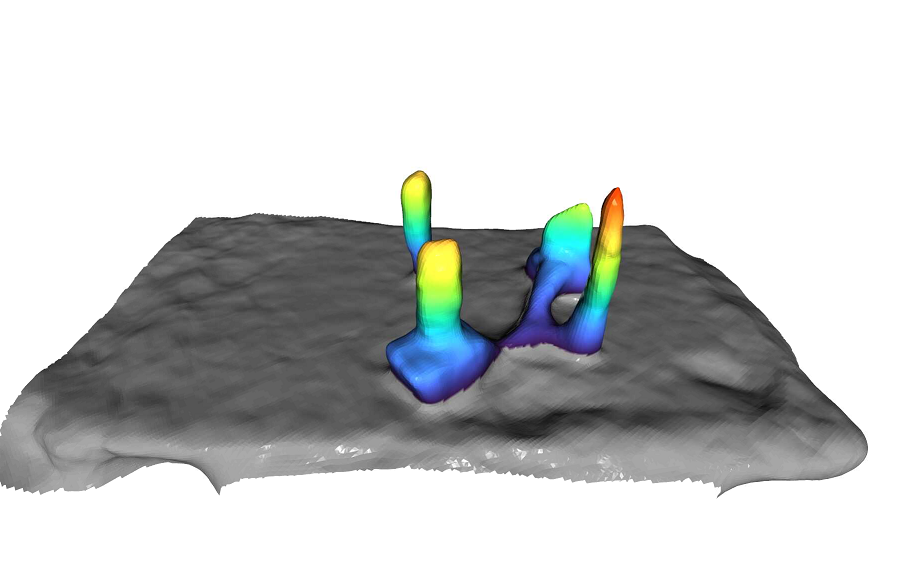} 
    \hspace{-1.9ex}
    &
    \hspace{\figzerolevelmeshspace}
    \includegraphics[height=\figzerolevelmeshscale]{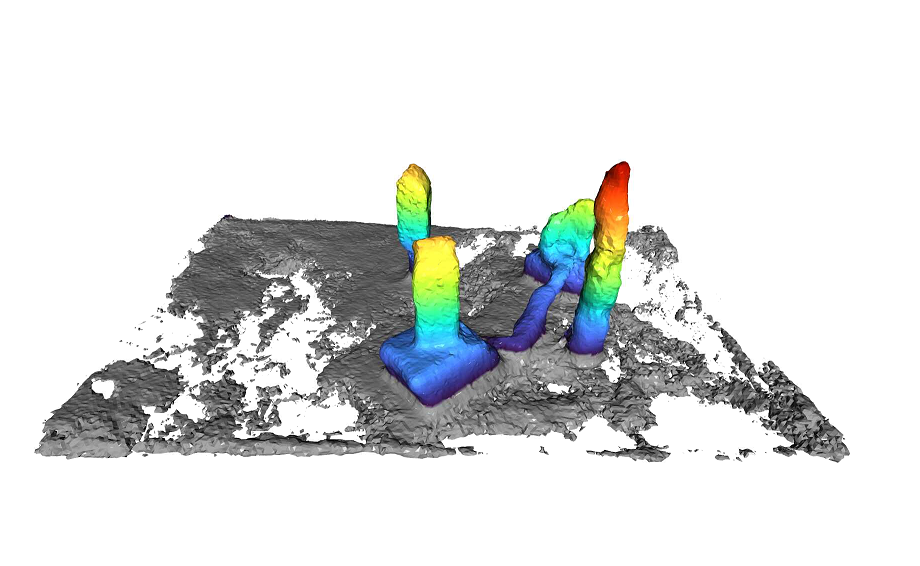}
    \hspace{\figzerolevelmeshspace}
\\
    \includegraphics[scale=0.24]{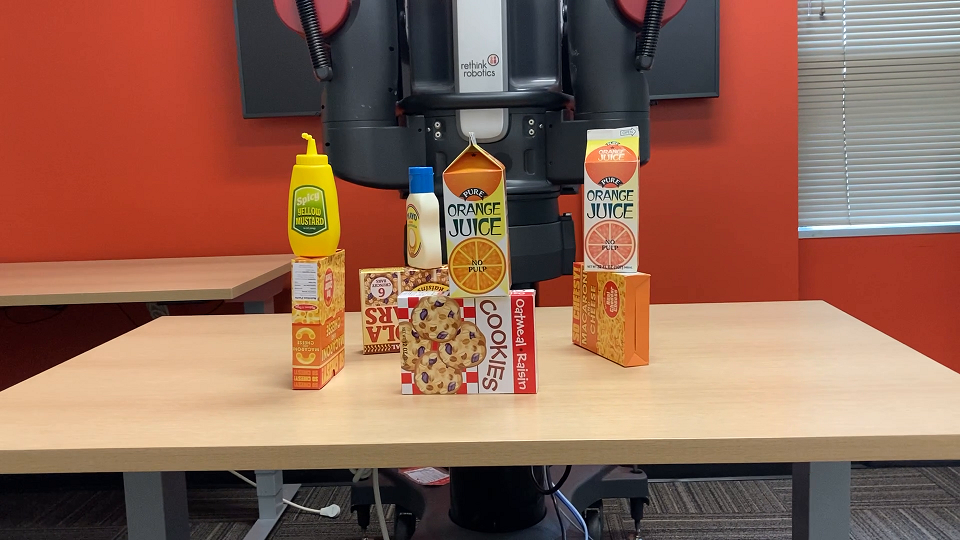} 
    &
    \hspace{\figzerolevelmeshspace}
    \includegraphics[height=\figzerolevelmeshscale]{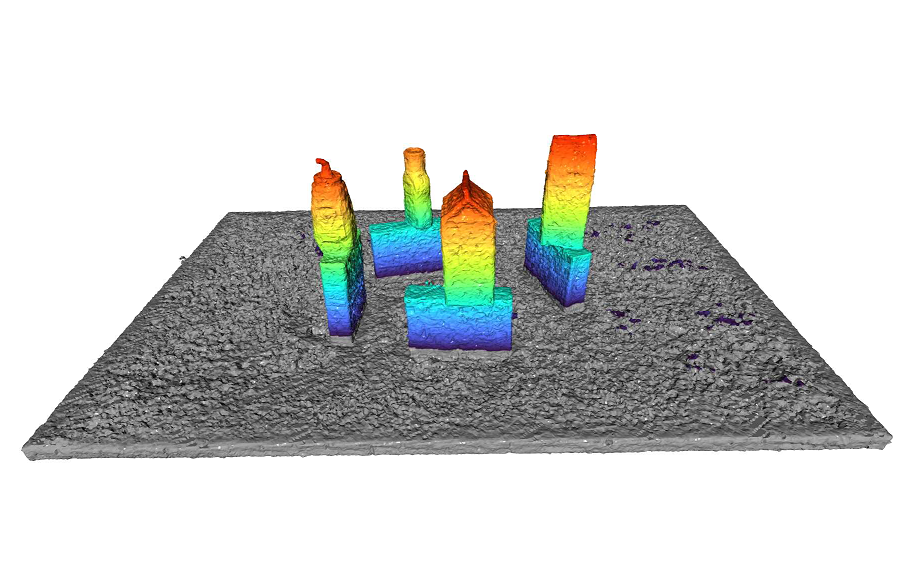} 
    \hspace{\figzerolevelmeshspace}
    &
    \hspace{\figzerolevelmeshspace}
    \includegraphics[height=\figzerolevelmeshscale]{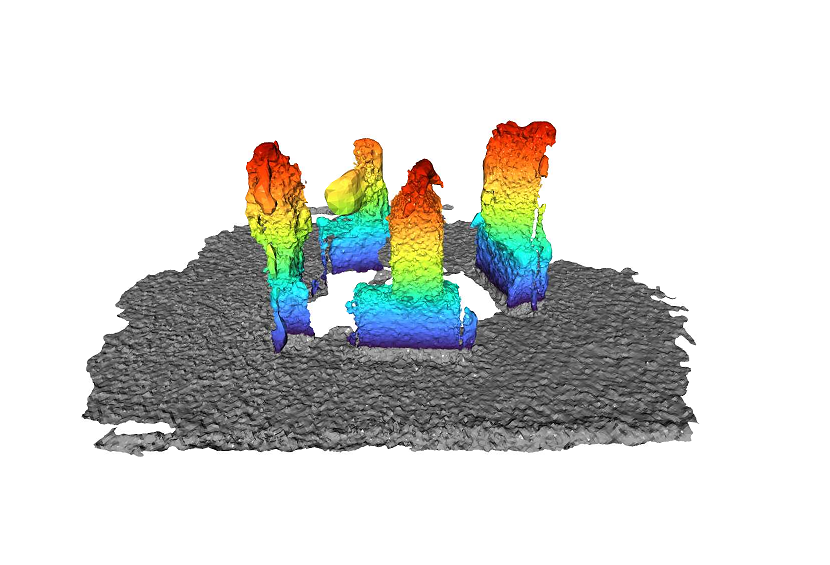} 
    \hspace{\figzerolevelmeshspace}
    &
    \hspace{-1.8ex}
    \includegraphics[height=\figzerolevelmeshscale]{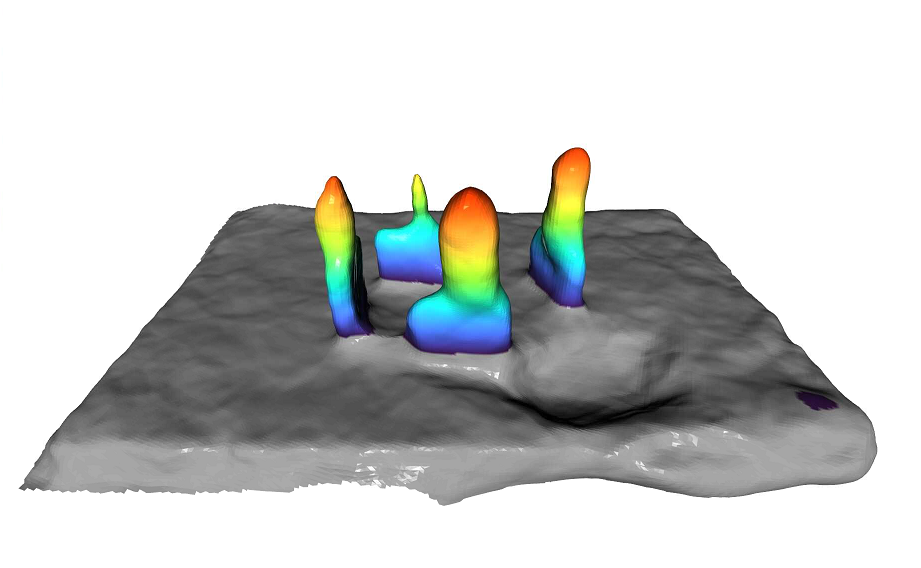} 
    \hspace{-1.9ex}
    &
    \hspace{\figzerolevelmeshspace}
    \includegraphics[height=\figzerolevelmeshscale]{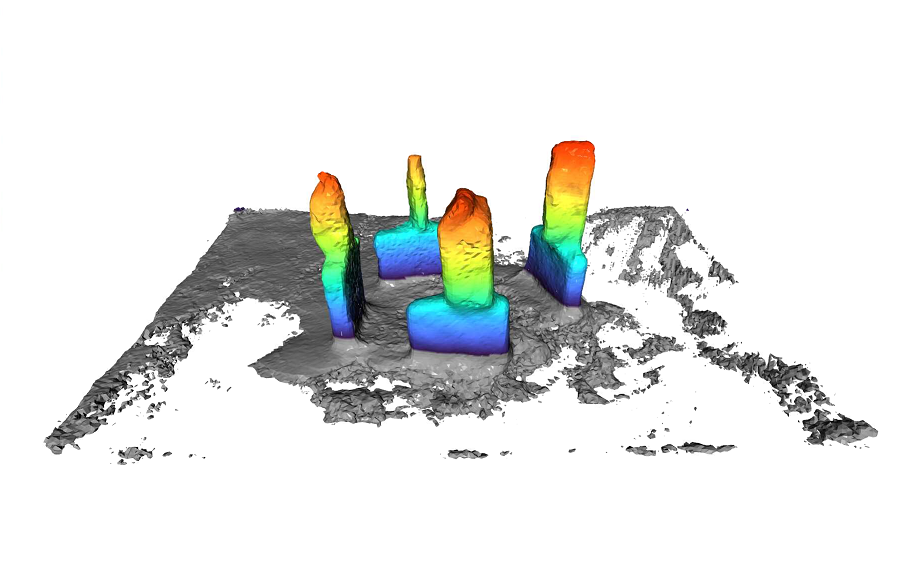} 
    \hspace{\figzerolevelmeshspace}
    \\
    One image of the scene & Ours & COLMAP & iSDF & Voxblox
  \end{tabular}
  \caption{{\sc Left:}  An image from each of the scenes.  {\sc Remaining columns:}  The mesh created by the zero level set of the SDF computed by several methods. Object meshes are colored by height using Turbo~\cite{mikhailov2019:turbo}.  Our method reconstructs objects accurately, while other methods struggle to fill holes and/or capture fine details. Average number of vertices/triangles among all scenes: 1.9M/3.8M (Ours), 240K/470K (COLMAP), 64K/130K (iSDF), 130K/190K (Voxblox).  }
  \label{fig:zerolevelmesh}
\end{figure*}

\begin{figure*}
   \def\figcloseupmeshscale{0.12}
    \centering
  \begin{tabular}{ccccc}
    \includegraphics[scale=\figcloseupmeshscale]{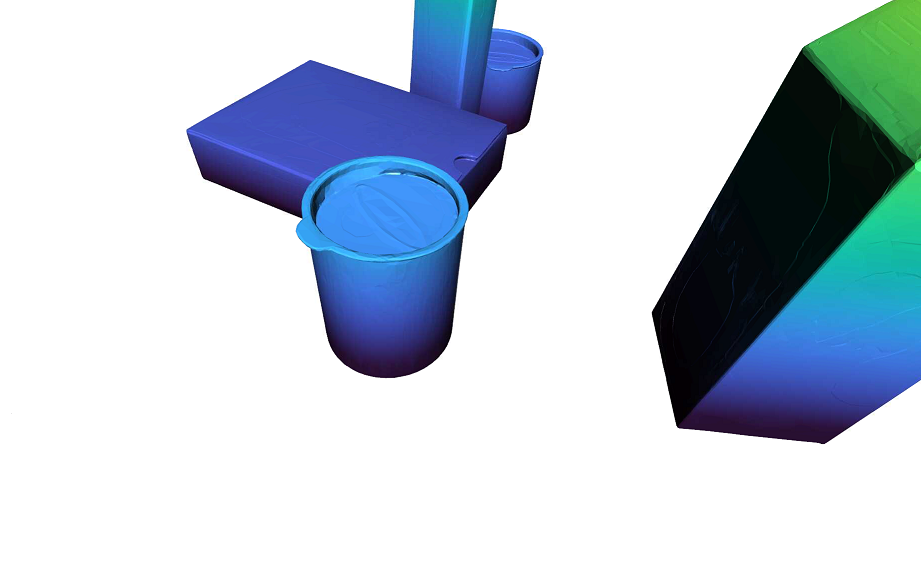} &
    \includegraphics[scale=\figcloseupmeshscale]{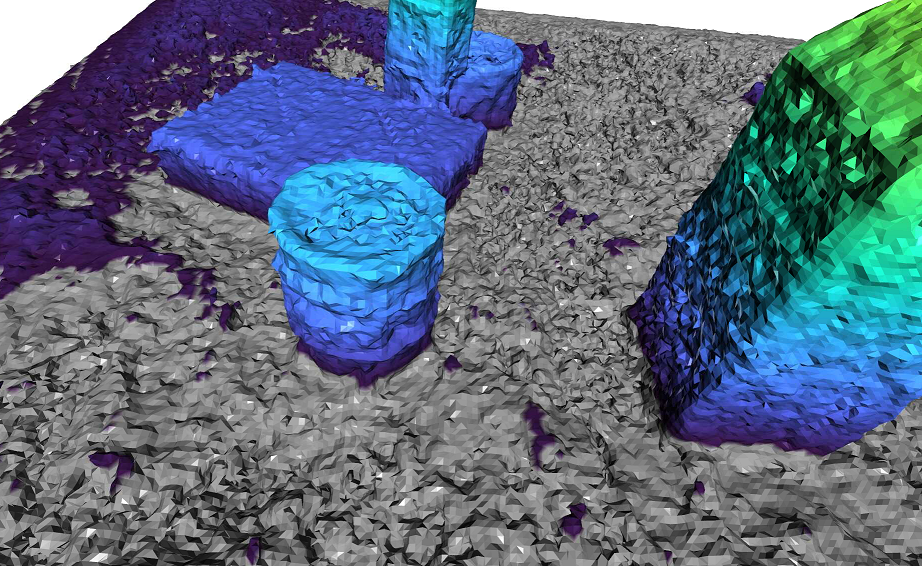} &
    \includegraphics[scale=\figcloseupmeshscale]{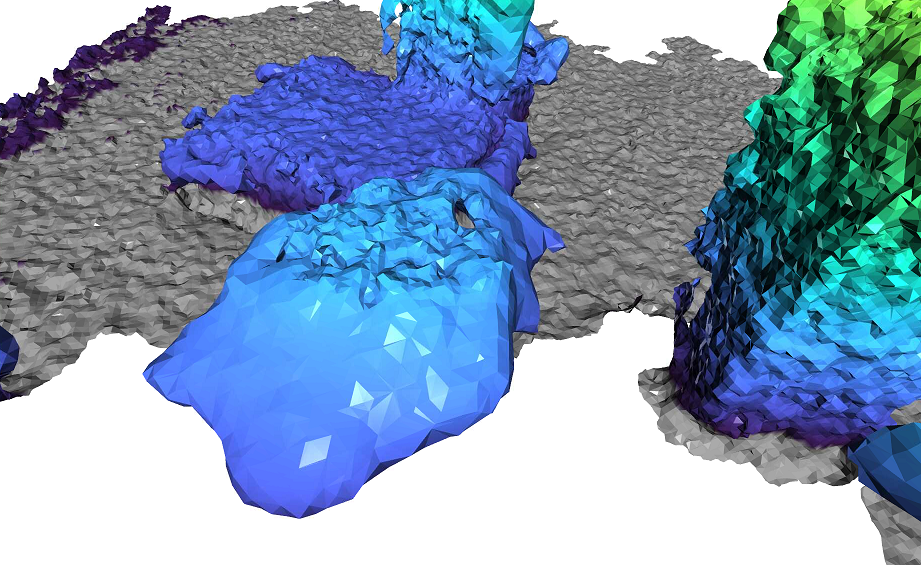} &
    \includegraphics[scale=\figcloseupmeshscale]{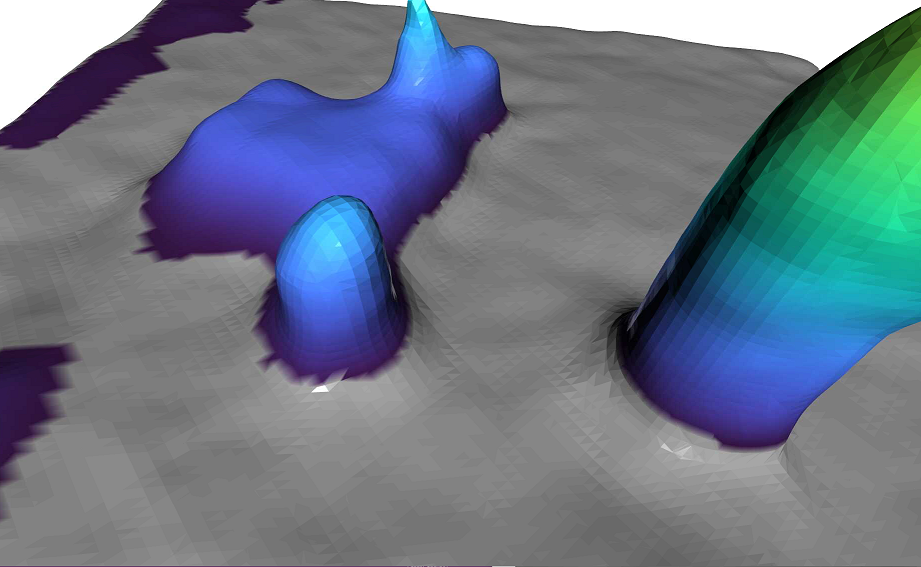} &
    \includegraphics[scale=\figcloseupmeshscale]{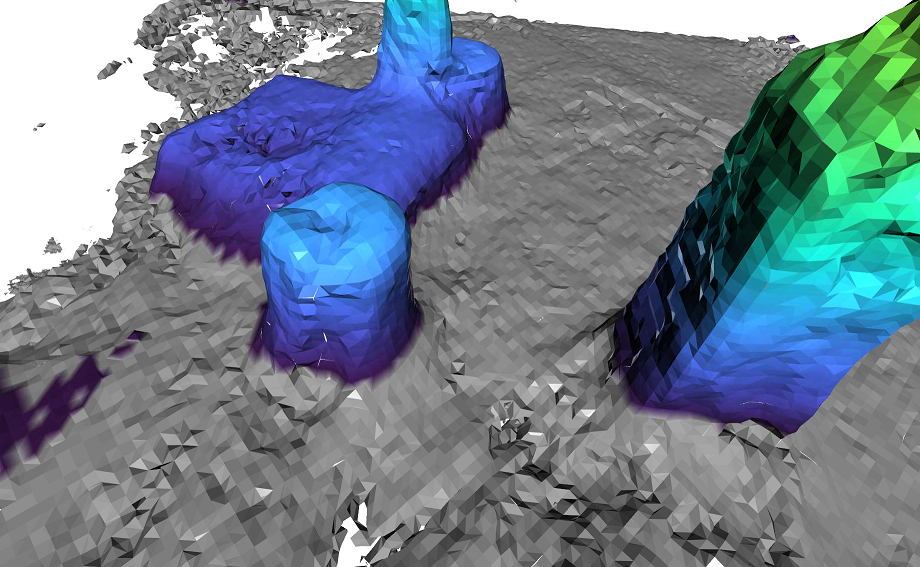}
    \\
    \includegraphics[scale=\figcloseupmeshscale]{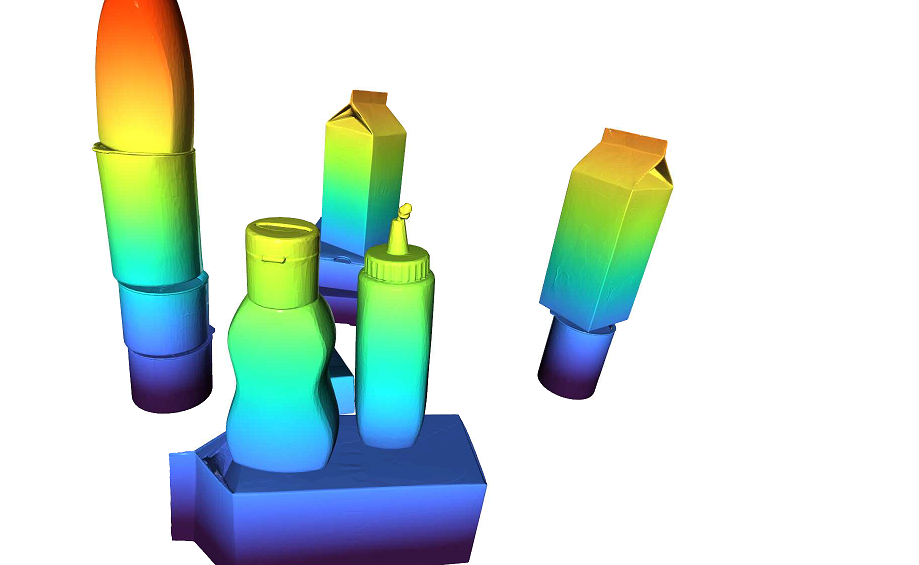} &
    \includegraphics[scale=\figcloseupmeshscale]{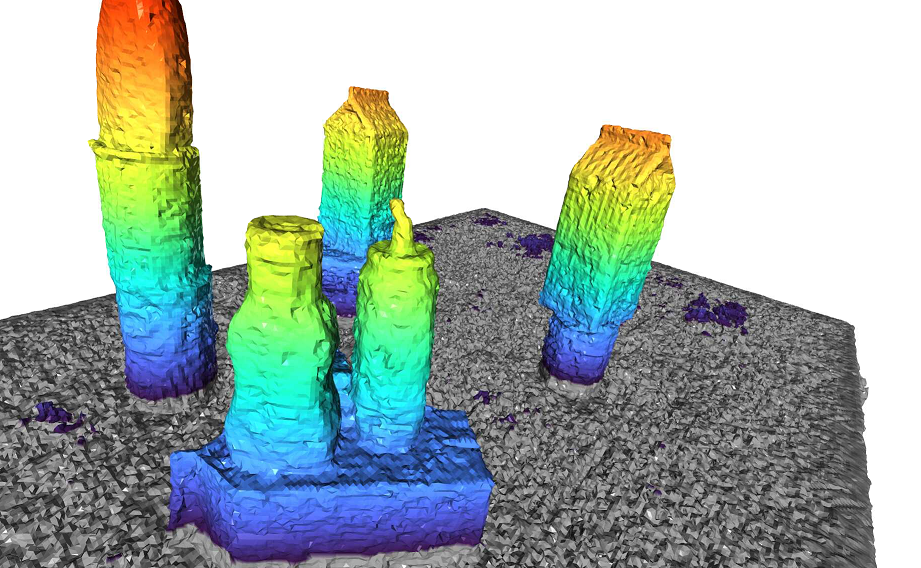} &
    \includegraphics[scale=\figcloseupmeshscale]{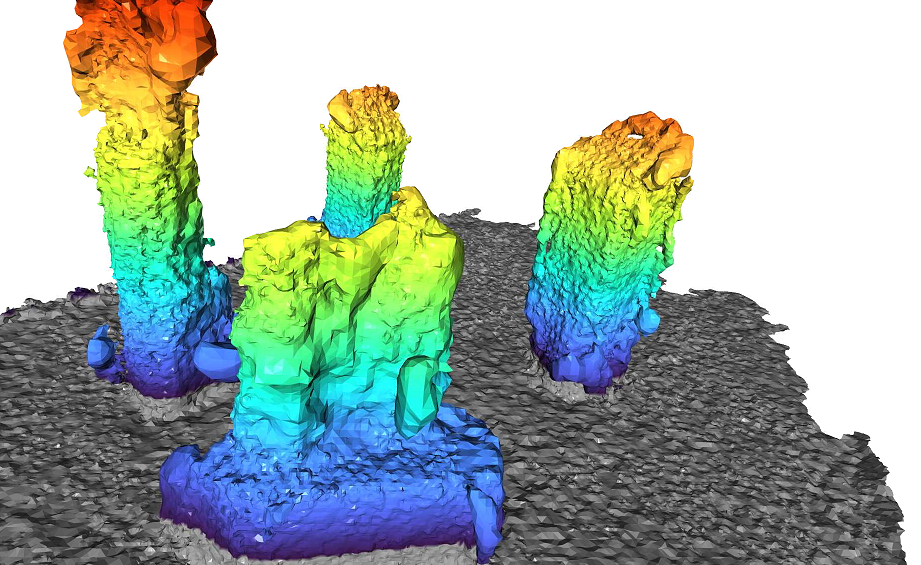} &
    \includegraphics[scale=\figcloseupmeshscale]{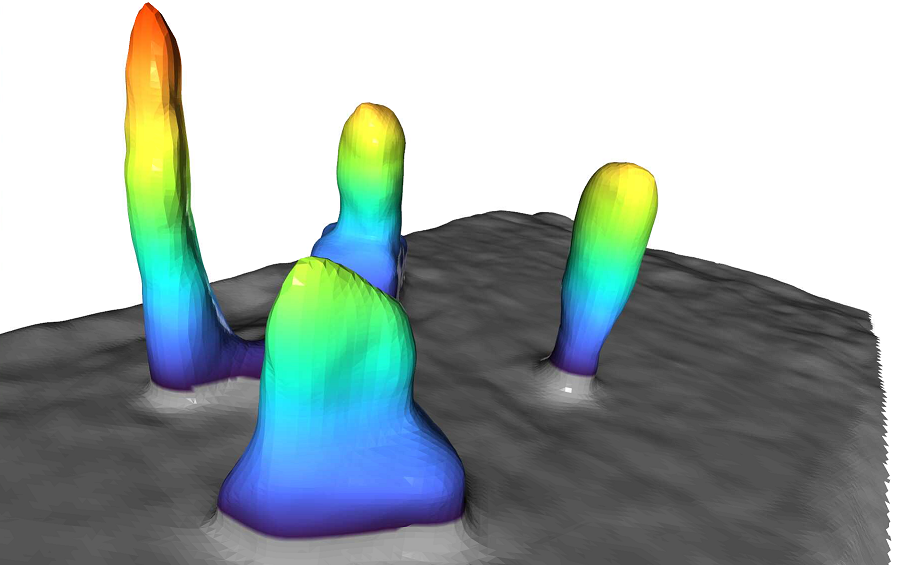} &
    \includegraphics[scale=\figcloseupmeshscale]{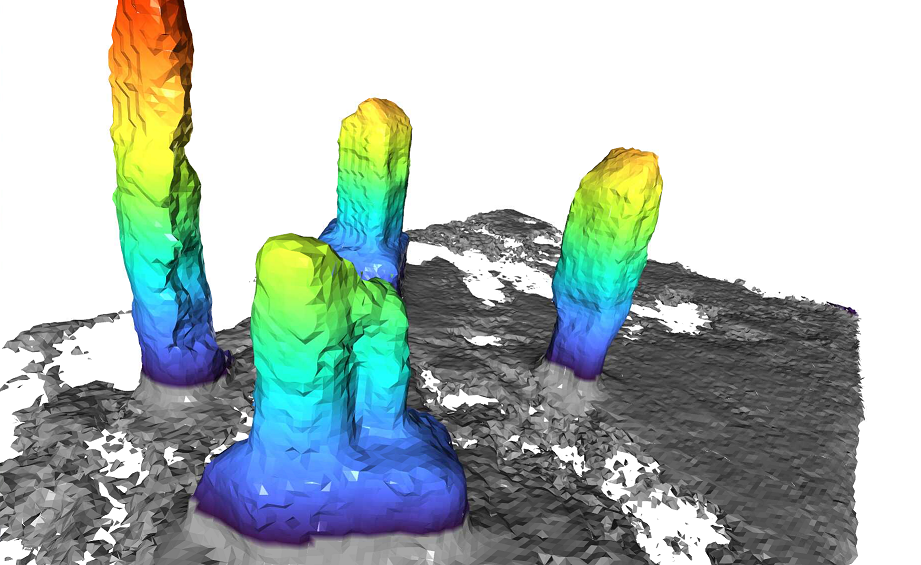} \\
    Ground truth & Ours & COLMAP & iSDF & Voxblox
    \end{tabular}
  \caption{Close-up of the mesh from the previous figure.
  {\sc Top row:}  Mesh from Scene~1, showing a can that is not fully observed due to partial occlusions from surrounding objects. COLMAP fails to build the can, and iSDF removes details.  {\sc Bottom row:}  Mesh from Scene~2, showing two  bottles in close proximity. Only our method is able to properly reconstruct the space between the bottles. Note that ground truth visualization omits the table.}
  \label{fig:closeup}
\end{figure*}

\subsection{Neural scene reconstruction from RGB images}
\label{sec:rgbngp}

Existing methods often rely on depth sensors to reconstruct the scene-level mesh for robotic planning and control~\cite{wen2022you}. There have been extensive studies on achieving this, such as using a depth-based SLAM~\cite{izadi2011kinectfusion,oleynikova2017iros:voxblox}, or direct voxelization \cite{hornung2013octomap,wen2022catgrasp}. However, when the depth modality is unavailable, which is the setup considered in this work, RGB-only scene-level reconstruction becomes more challenging. While multi-view stereo based methods \cite{schoenberger2016sfm} are able to triangulate 3D scene points from 2D feature matching, the reconstructed point clouds are sparse, especially for textureless regions. Meshing from those point clouds inevitably leads to holes or non-manifold topology, which are not compatible with motion planning algorithms (see experiments in  Sec.~\ref{sec:exp}), %
which require a continuous and differential representation of the scene.  To overcome this limitation, in this work we leverage the recent advances in neural radiance fields (NeRFs) to learn a compact, accurate and holistic representation of the scene in an online fashion. For the sake of fast training and instant application for downstream robotic tasks, we adopt neural graphics primitives (NGP) \cite{mueller2022siggraph:instantngp}, a much faster alternative to NeRF \cite{mildenhall2020eccv:nerf} based on grid feature encoding via voxel hashing.  
Learning a new scene with instant NGP~\cite{mueller2022siggraph:instantngp} can be done in a few seconds.

Concretely, the neural radiance field takes as input a  query 3D position $x\in \mathbb{R}^3$ and a 3D viewing direction $d\in \mathbb{R}^3, \|d\|=1$. The output is the RGB value $c\in [0,1]^3$ and the density value $\sigma \in [0,\infty)$. This can be written as 
$f(x,d)\mapsto(c,\sigma)$, where $\sigma$ indicates the differential likelihood of a ray hitting a particle (i.e., the probability of hitting a particle while traveling an infinitesimal distance)~\cite{mildenhall2020eccv:nerf}. Given the multi-view RGB images with known camera poses obtained using SLAM methods~\cite{schoenberger2016cvpr:sfm,schoenberger2016eccv:mvs},  query points are allocated by sampling various traveling times $t$ along the ray  $r(t)=o_w+t \cdot d_w$, where $o_w$ and $d_w$ denote camera origin and ray direction in the world frame, respectively. Based on volume rendering, the final color of the ray is then integrated via alpha compositing:
\begin{gather}
\hat c(r)=\int_{t_n}^{t_f}T(t)\sigma(r(t))c(r(t),d) \, dt \\
T(t)=\exp\left(-\int_{t_n}^{t}\sigma(r(s))\,ds\right).
\end{gather}

In practice, the integral is approximated by quadrature. Following \cite{mueller2022siggraph:instantngp}, the neural radiance field function $f(\cdot)=\textbf{\text{MLP}}(\textbf{\text{enc}}(\cdot))$ is composed of a multi-scale grid feature encoder $\textbf{\text{enc}}$ and a multilayer perceptron (\textbf{MLP}). $f$ is optimized per-scene by minimizing the $L_2$ loss between volume rendering and the corresponding pixel value obtained from RGB images, i.e., by minimizing
\begin{gather}
\mathcal{L}=\sum_{r\in R}^{}\left \| \bar{c}(r)-\hat c(r) \right \|_2.
\end{gather}
Here, $\hat c$ is the predicted integral RGB value along the ray; $\bar{c}$ is the true observation from the RGB image at the pixel location through which the ray $r$ travels; and $R$ is the total ray set for training. Once trained, we can query the density values on a dense grid of point locations. This permits the use of Marching Cubes~\cite{lorensen1987siggraph:marchcube} to extract a polygonal mesh capturing an isosurface from the density field~\cite{mueller2022siggraph:instantngp}.  
A bounding volume hierarchy is created from the mesh, which is then queried for signed distance values.

\subsection{Manipulator control problem}
\label{sec:mpc}
We leverage the sampling-based model predictive algorithm developed by Bhardwaj~et al.~\cite{bhardwaj2021corl:storm}, extended to support mesh-based signed distance function in Sundaralingam~et al.~\cite{balakumar2023curobo}. We then load the reconstructed mesh of the scene from our approach into this model predictive controller for collision-free motions of the robot.

\begin{figure*}
   \def\figsdfscale{0.29}
    \centering
    \begin{tabular}{cccccc}
    \!\includegraphics[scale=\figsdfscale]{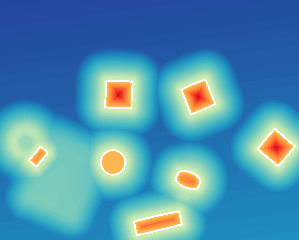} & 
    \!\includegraphics[scale=\figsdfscale]{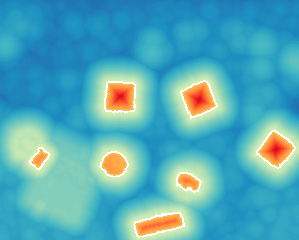} &
    \!\includegraphics[scale=\figsdfscale]{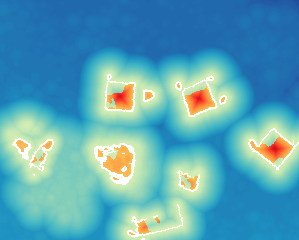} &
    \!\includegraphics[scale=\figsdfscale]{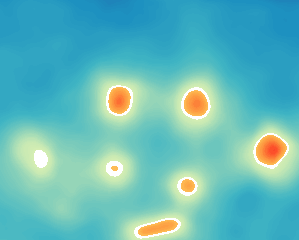} & 
    \!\includegraphics[scale=\figsdfscale]{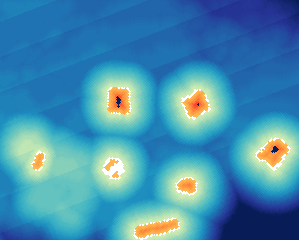} &
    \multirow{3}{*}{\!\includegraphics[scale=0.4]{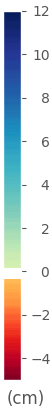}}
    \\
    \!\includegraphics[scale=\figsdfscale]{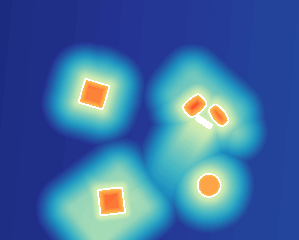} &
    \!\includegraphics[scale=\figsdfscale]{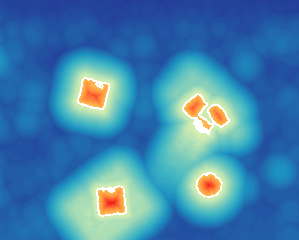} &
    \!\includegraphics[scale=\figsdfscale]{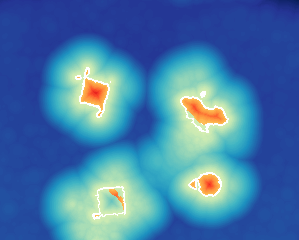} &
    \!\includegraphics[scale=\figsdfscale]{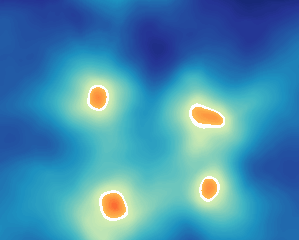} &
    \!\includegraphics[scale=\figsdfscale]{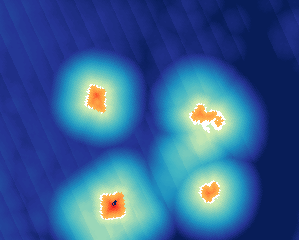}
    \\
    \!\includegraphics[scale=\figsdfscale]{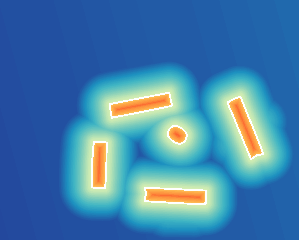} &
    \!\includegraphics[scale=\figsdfscale]{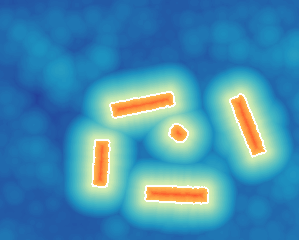} &
    \!\includegraphics[scale=\figsdfscale]{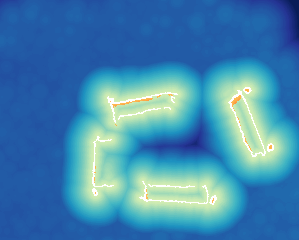} &
    \!\includegraphics[scale=\figsdfscale]{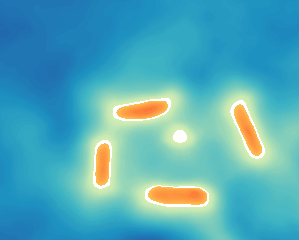} &
    \!\includegraphics[scale=\figsdfscale]{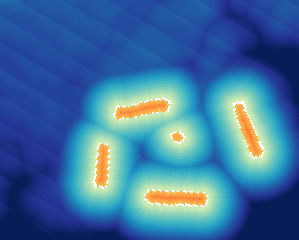}
    \\
    Ground truth & Ours & COLMAP & iSDF & Voxblox
    \end{tabular}
  \caption{Comparison of SDF computed by several methods.   {\sc Top to bottom:}  Scene 1, Scene 2, Scene 3.  Shown is a 2D horizontal slice of the SDFs in Fig.~\ref{fig:zerolevelmesh} approximately 7~cm above the table.  Our method is more accurate, and it generates water-tight meshes. Unlike COLMAP, which is conservative, both iSDF and Voxblox shrink the objects, leading to collision (see Tab.~\ref{tab:errorcomp2}). Note also that COLMAP's meshes are not water-tight. %
}
  \label{fig:sdfslice}
\end{figure*}

\begin{figure*}
   \def\figcloseupmeshscale{0.13}
    \centering
  \begin{tabular}{cccc}
    \!\!\raisebox{7ex}{\rotatebox{90}{Ours}}\!\!\!\! & \includegraphics[scale=\figcloseupmeshscale]{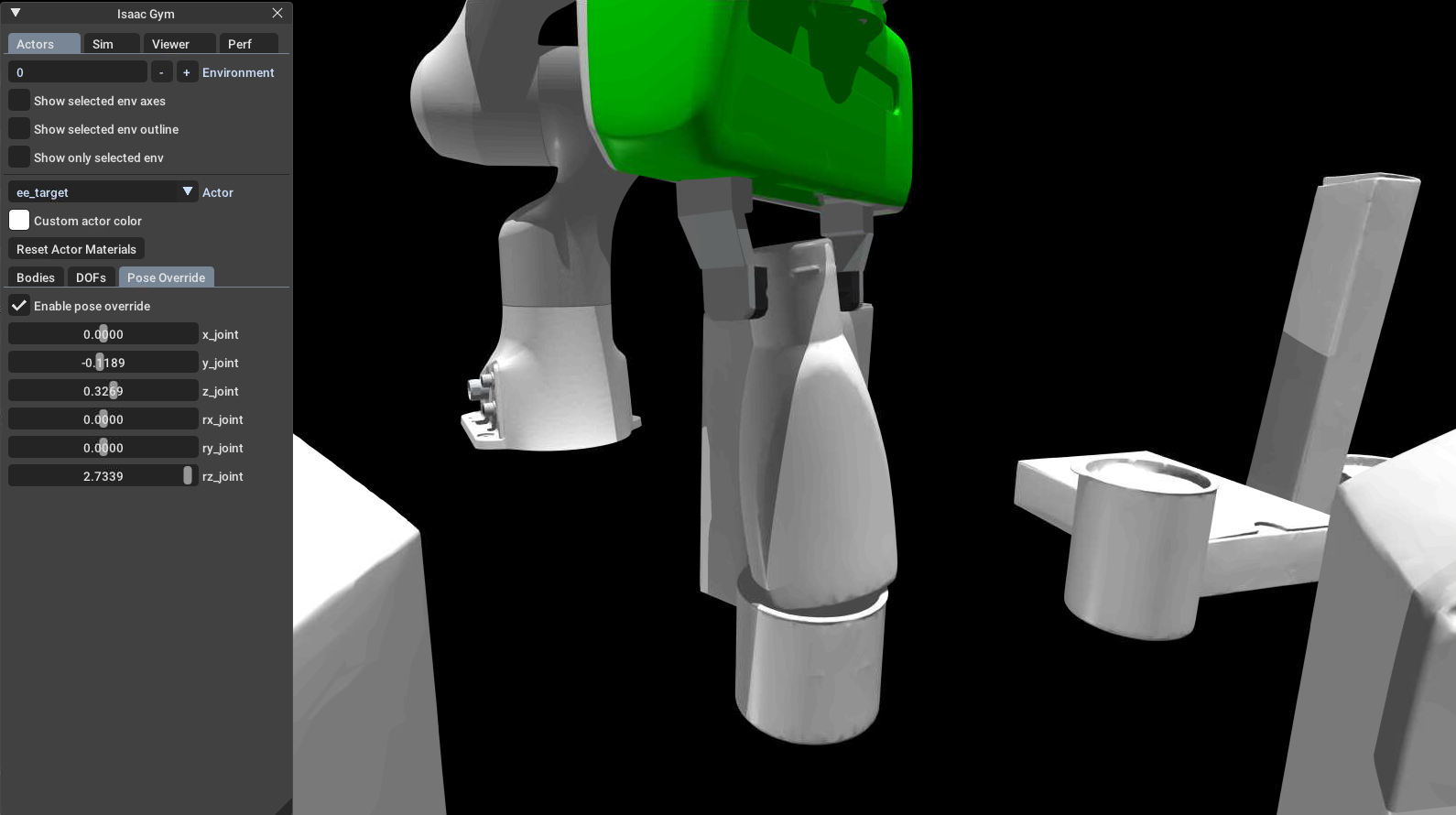} &
    \hspace{-1.8ex}\includegraphics[scale=\figcloseupmeshscale]{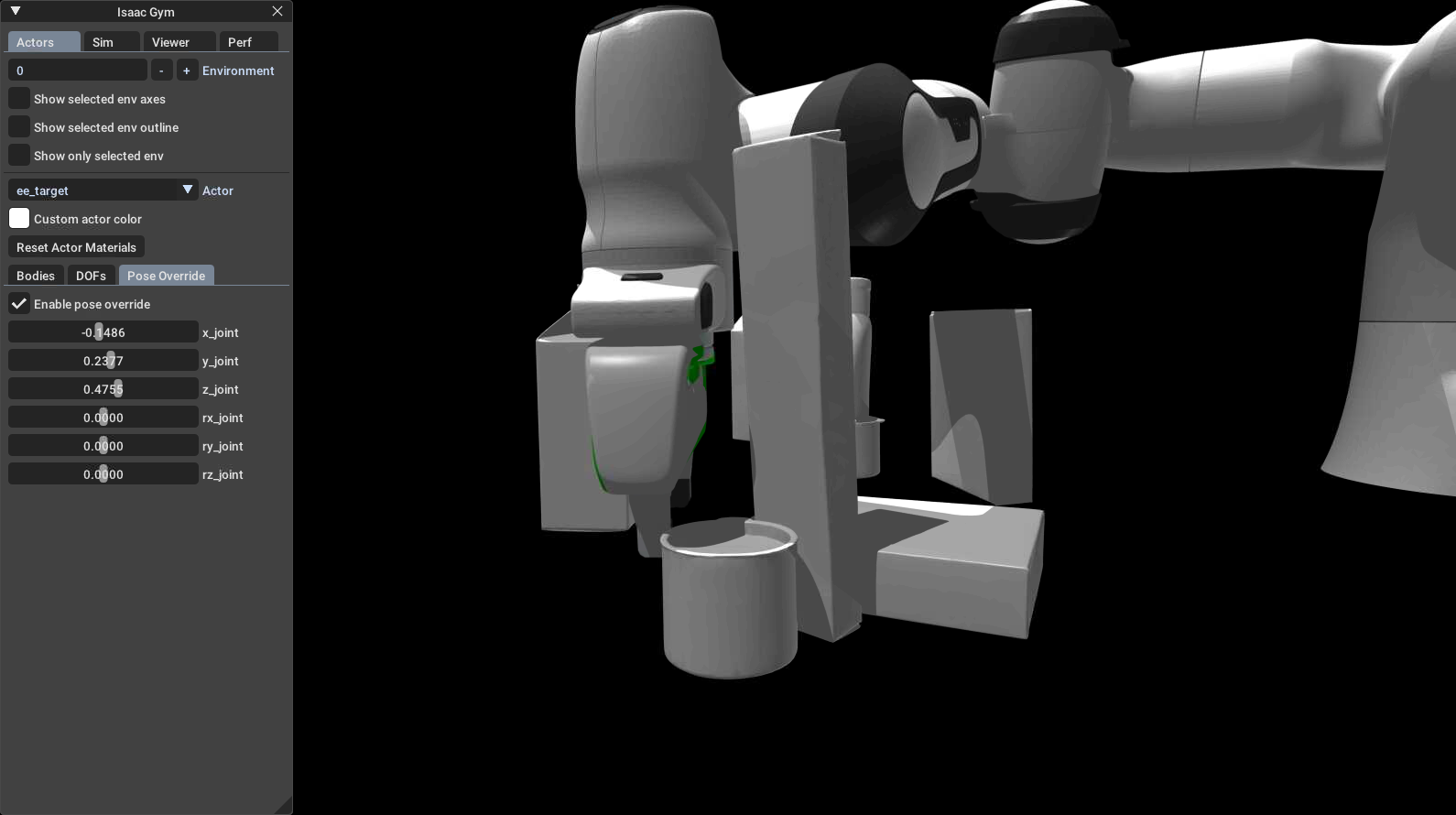} &
    \hspace{-1.8ex}\includegraphics[scale=\figcloseupmeshscale]{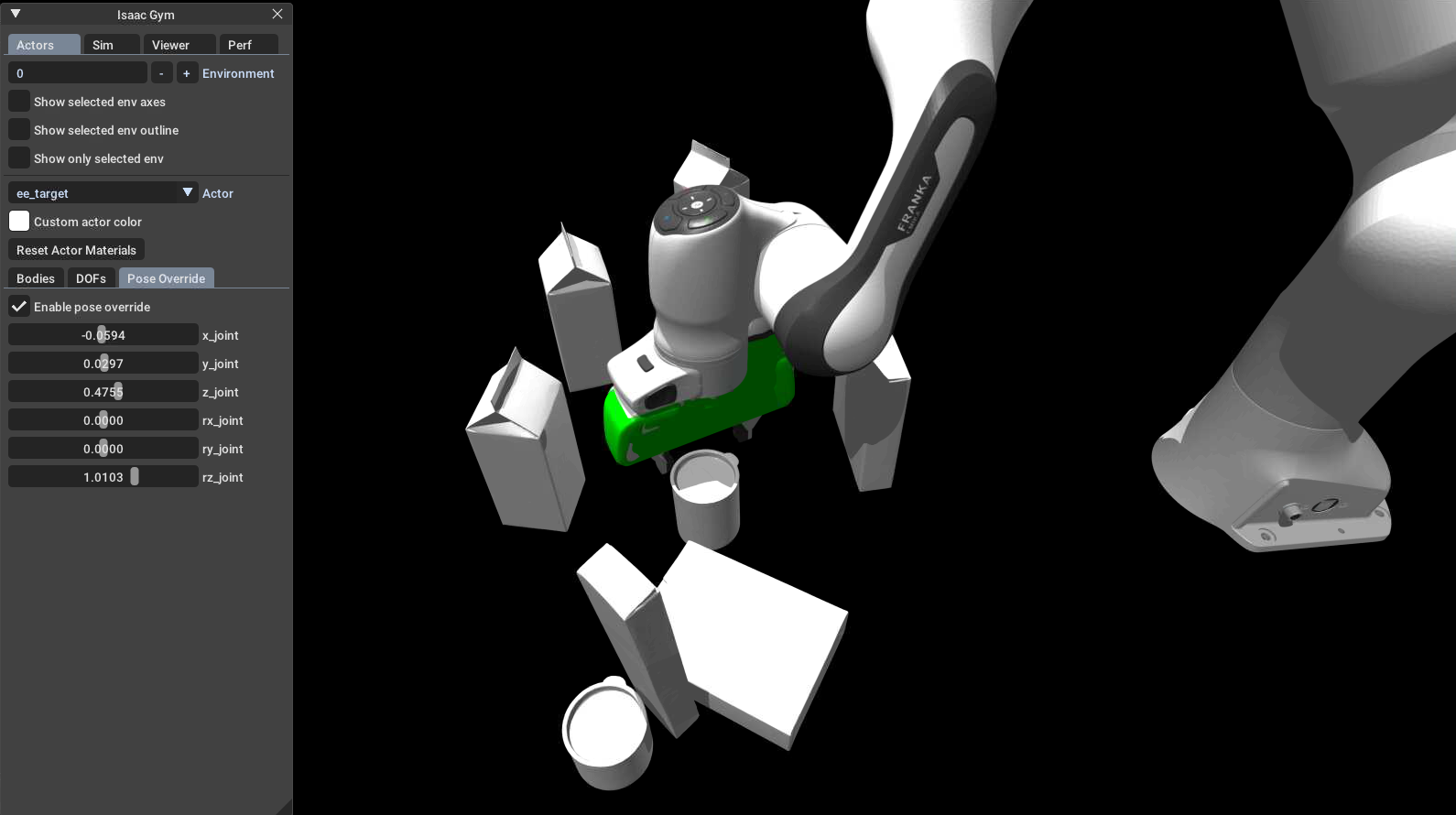}
    \\
    \!\!\raisebox{5ex}{\rotatebox{90}{COLMAP}}\!\!\!\! & \includegraphics[scale=\figcloseupmeshscale]{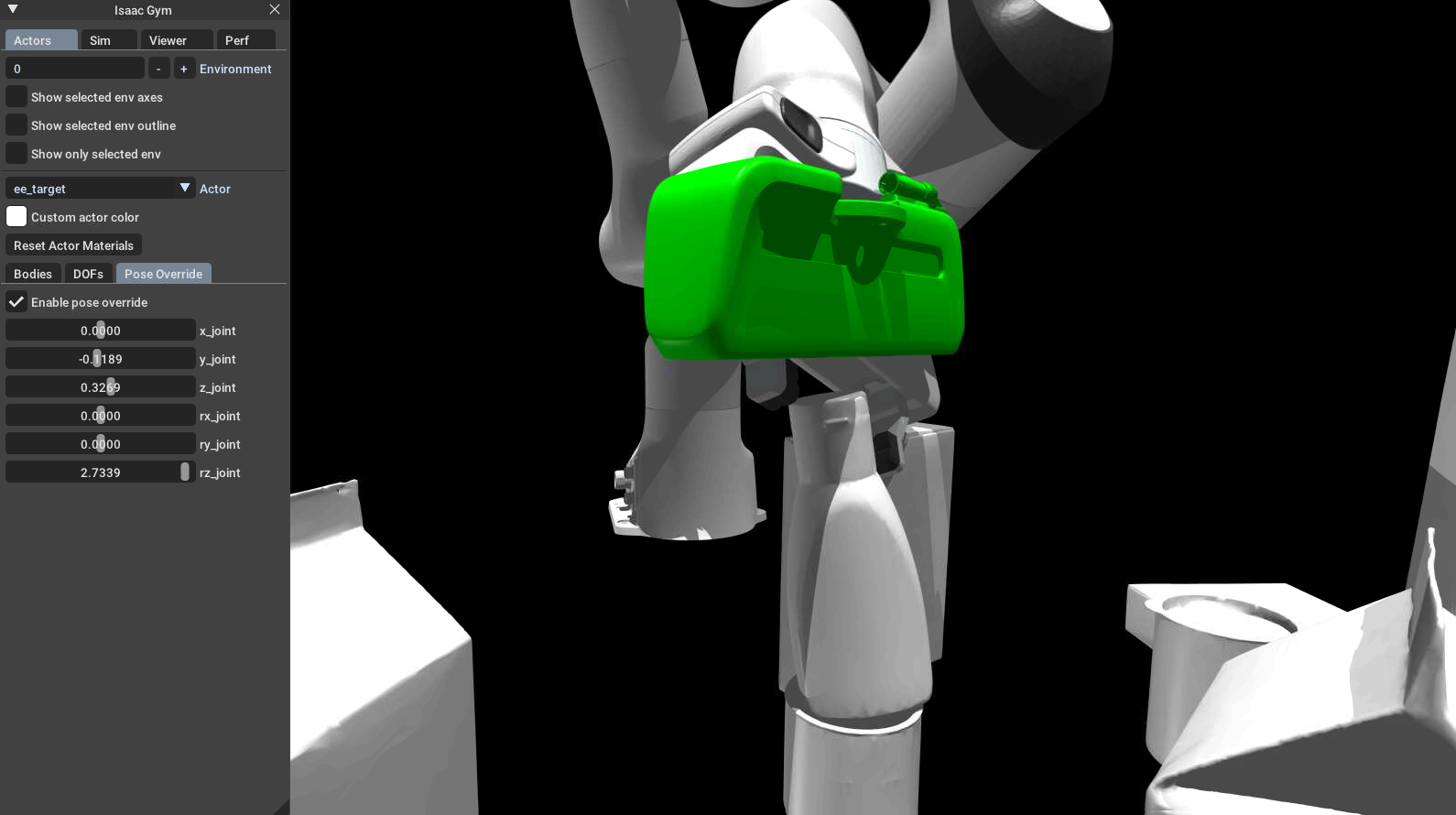} &
    \hspace{-1.8ex}\includegraphics[scale=\figcloseupmeshscale]{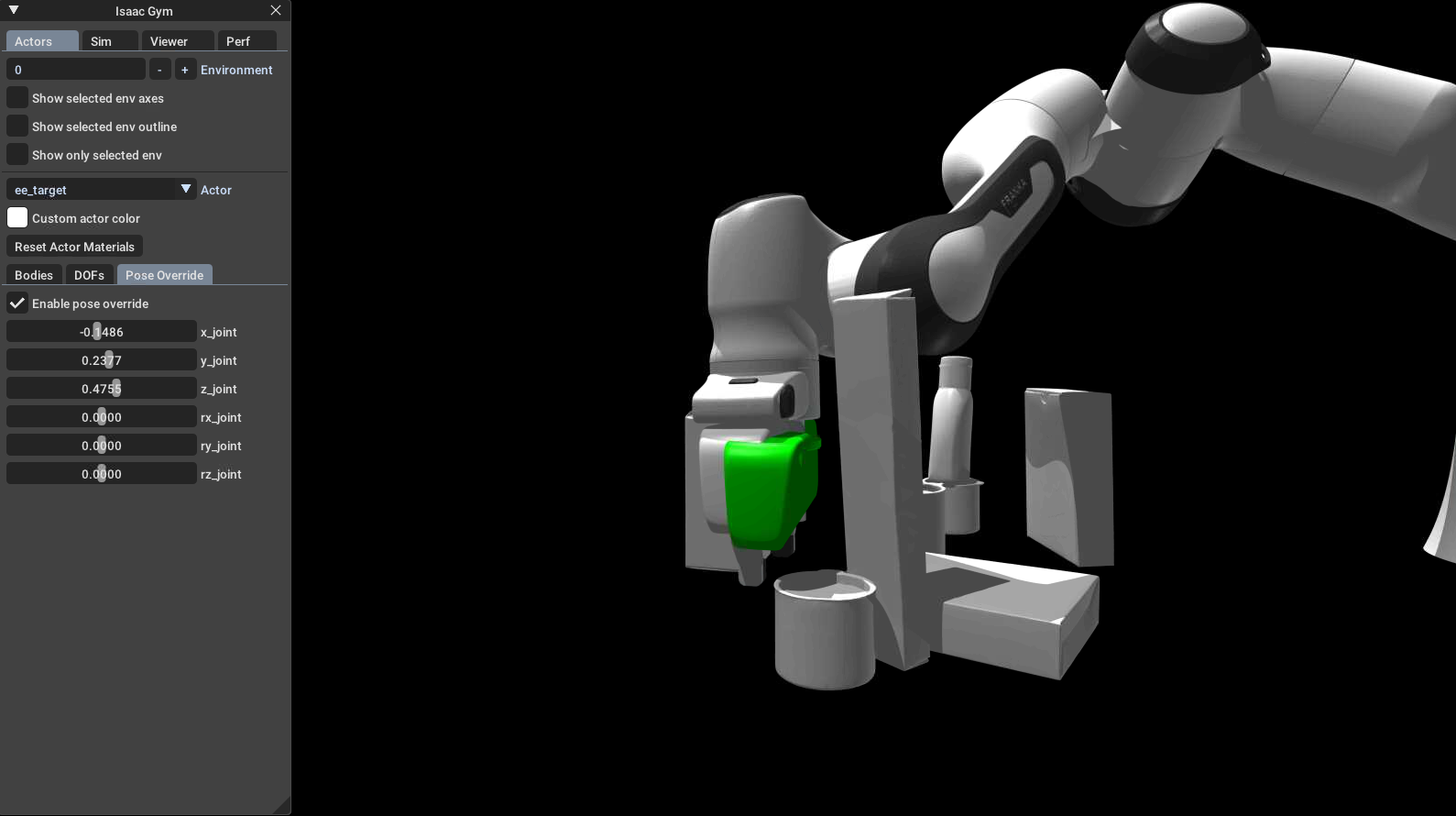} &
    \hspace{-1.8ex}\includegraphics[scale=\figcloseupmeshscale]{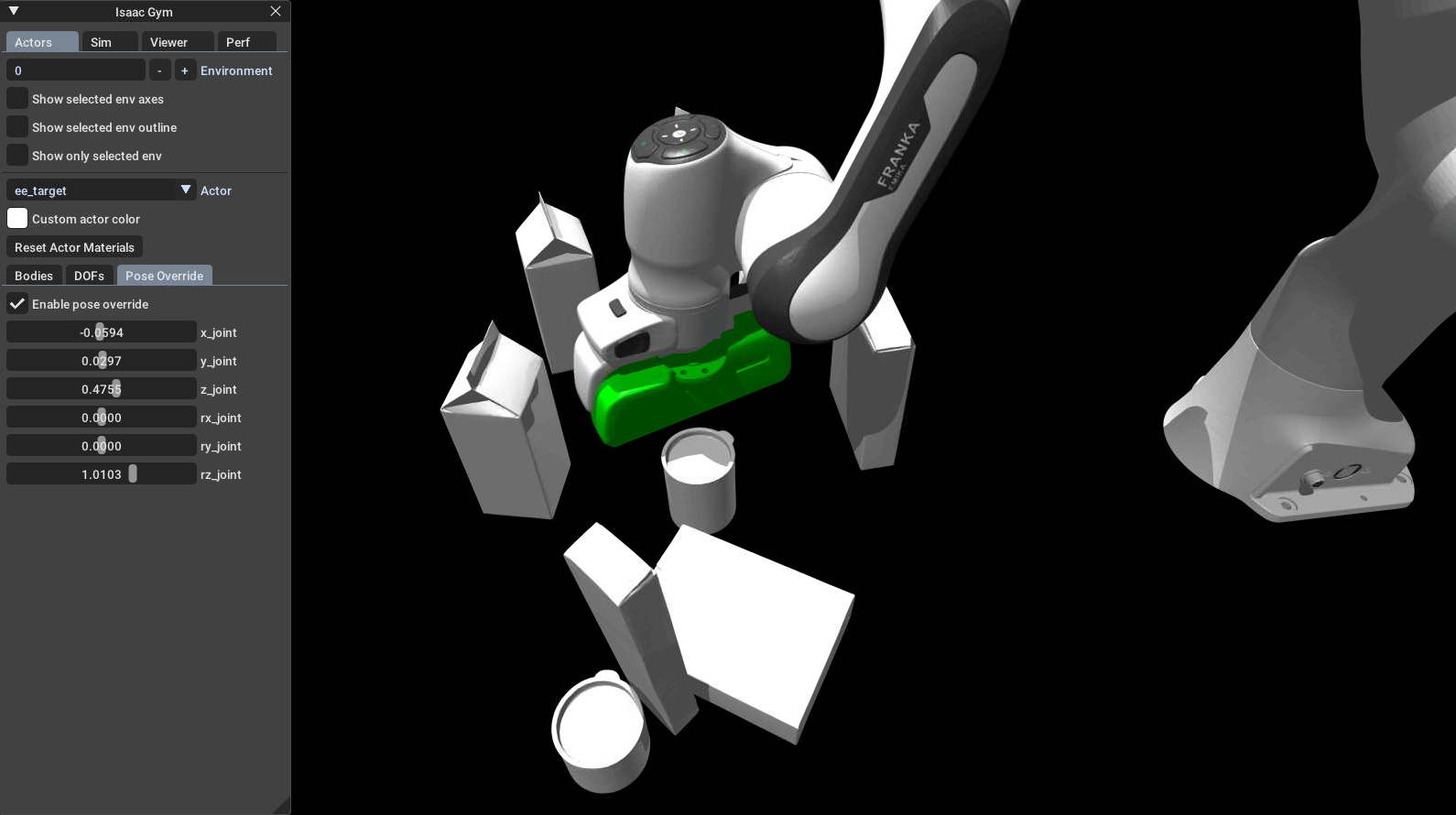}
    \end{tabular}
  \caption{Comparison of our method (first row) to COLMAP reconstruction (second row). The robot manipulator is asked to move to a target pose (indicated by the green gripper) for three scenarios (left to right).  {\sc Top:}  Our method reaches the target with high accuracy.  In the middle column, the accuracy is so high that the green target is almost perfectly occluded by the gripper.  {\sc Bottom:}  Because of its conservative SDF estimation, COLMAP is not able to reach the target when there is significant clutter.}
  \label{fig:gripper}
\end{figure*}

\section{EXPERIMENTAL RESULTS}
\label{sec:exp}

To evaluate our method, we run experiments on real image data.
We divide these experiments into two parts.
First we measure the geometric accuracy of 3D reconstruction of the perception system.
Then we evaluate the entire perception-and-control system by measuring the maximum penetration, the error at the goal, and the success rate.

\subsection{Real tabletop dataset with ground truth}

We created three real-world tabletop scenes in front of a robot manipulator.
Fig.~\ref{fig:tabletopx} shows images from the three scenes.
The objects on the table were taken from the HOPE~\cite{tyree2022iros:hope} dataset to facilitate ground truth mesh creation.
For each scene, we placed approximately ten objects on the table, and we captured the static scene twice with different sensors.
One sequence was captured with a RealSense D415 RGBD sensor at a resolution of $640 \times 480$ for both RGB and depth.
The other RGB-only sequence was captured with an iPhone SE rear-facing sensor at full HD resolution ($1920 \times 1080$) following a similar capture trajectory.
Each sequence contained approximately 3000 frames, and we automatically selected the sharpest frame from each subsequence of ten consecutive frames, leading to approximately 300 sharp image frames.

The ground truth camera poses were obtained by running the structure-from-motion algorithm in COLMAP~\cite{schoenberger2016cvpr:sfm,schoenberger2016eccv:mvs} on each sequence.
The ground truth mesh was obtained by running multi-view CosyPose~\cite{labbe2020eccv:cosypose} on each sequence to recover the 6-DoF (degree-of-freedom) poses of the objects.
From these poses, the relative rotation, translation, and scale between the two sequences was estimated via Procrustes analysis.
A plane was fit to the table, and the HOPE meshes, along with the table plane, were used to generate the ground truth mesh for each scene.

\subsection{ESDF reconstruction}

We compare our method with three leading techniques:  the multi-view reconstruction method in COLMAP~\cite{schoenberger2016cvpr:sfm,schoenberger2016eccv:mvs}, iSDF~\cite{ortiz2022rss:isdf}, and Voxblox~\cite{oleynikova2017iros:voxblox}.
iSDF takes as input a set of posed depth images, from which it learns a neural representation of the scene as a Euclidean full SDF that can be queried at any 3D point.
COLMAP not only computes the camera poses via structure-from-motion, but it also produces a 3D point cloud representation of the scene via multi-view stereo at discrete feature locations.
From this point cloud, a mesh is generated via Poisson surface reconstruction.
Voxblox also receives posed depth images and build the Euclidean full SDF incrementally, but it stores SDF on voxels rather than using a network representation. For fair comparison, we increased the voxel resolution of Voxblox to 0.5~cm, we ran COLMAP with three different image resolutions (by downsampling the horizontal dimension to 250, 500, and 1000 pixels), and we ran our method with four different iteration lengths (1k, 2k, 4k, and 8k).  On a PC with NVIDIA GeForce RTX 3090 GPU, the resulting runtimes were 72~s for our method (with 8k iterations), 78~s for COLMAP (with small image resolution), 90~s for Voxblox, and 20~min for iSDF.  (All runtimes exclude the structure-from-motion part of COLMAP that estimates camera poses, which is costly.)

Figs.~\ref{fig:zerolevelmesh} and \ref{fig:closeup} show the 3D reconstruction of the three scenes by our method and the three baselines. 
(We used 8k iterations for our method, and small image resolution for COLMAP.)
Our method more accurately captures the shape of the objects than the baselines.
A 2D horizontal slice through the resulting SDF is shown in Fig.~\ref{fig:sdfslice} for each scene and method, which again reveals the fidelity of our reconstruction.

\begin{figure}
   \def\plotthresscale{0.40}
    \centering
   \begin{tabular}{cc}
    \hspace{-2ex}\includegraphics[scale=\plotthresscale]{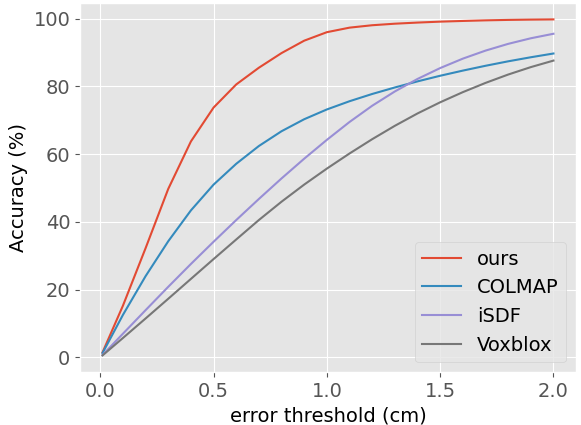} &
    \includegraphics[scale=\plotthresscale]{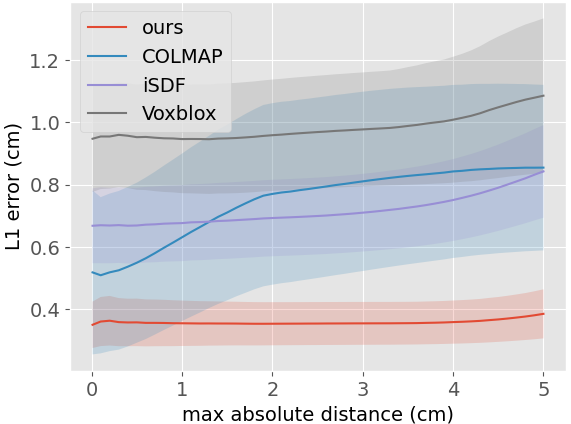}
  \end{tabular}
  \caption{{\sc Top:}  Accuracy of SDF estimation versus threshold, among all locations within 5~cm of the mesh surface.  With our method, approximately 95\% of the sampled locations have an error less than 1~cm. Curves include points from all three scenes.  {\sc Bottom:} The L1 error among all points within the specified distance of the mesh surface (horizontal axis). Shown are mean (solid lines) and 0.25x standard deviation (shaded areas), where the scaling factor on std affords better illustration.  Our method yields consistently low values for a wide variety of thresholds. Curves and shades represent the average value and standard deviation among three scenes.} 
  \label{fig:accplot}
\end{figure}

\begin{table*}
\centering
\caption{Error comparison over a regularly sampled volume of locations on the table.  Shown is mean~$\pm$~std.}
\begin{tabular}{cccccccccc}
\toprule
\hspace{-1em}Metric & \multicolumn{3}{c}{COLMAP~\cite{schoenberger2016cvpr:sfm,schoenberger2016eccv:mvs}} & iSDF & Voxblox & \multicolumn{4}{c}{Ours} \\
& small & med & large & \cite{ortiz2022rss:isdf} & \cite{oleynikova2017iros:voxblox} & 1k & 2k & 4k & 8k \\
\cmidrule(r){2-4}
\cmidrule(r){5-5}
\cmidrule(r){6-6}
\cmidrule(r){7-10}
collision ($\%$) $\downarrow$ & 48.5$\pm$17.9 & 40.1$\pm$20.9 & 38.1$\pm$20.5 & 34.7$\pm$7.3 & 35.7$\pm$9.0 & 19.0$\pm$18.5 & 6.9$\pm$3.0 & 6.6$\pm$3.5 & \textbf{6.5}$\pm$3.6\\
misclassification ($\%$) $\downarrow$ & 2.4$\pm$0.7 & 1.9$\pm$0.8 & 1.6$\pm$0.8 & 1.7$\pm$0.1 & 1.4$\pm$0.3 & 1.6$\pm$0.4 & 1.1$\pm$0.3 & \textbf{0.9}$\pm$0.2 & \textbf{0.9}$\pm$0.2\\
mean L1 (cm) $\downarrow$ & 1.9$\pm$0.3 & 1.6$\pm$0.3 & 1.5$\pm$0.4 & 1.1$\pm$0.2 & 1.5$\pm$0.3 & 1.0$\pm$0.5 & 0.5$\pm$0.2 & 0.5$\pm$0.1 & \textbf{0.4}$\pm$0.0\\
max L1 (cm) $\downarrow$ & 8.1$\pm$1.6 & 6.6$\pm$0.2 & 6.6$\pm$0.2 & 2.6$\pm$0.3 & 4.7$\pm$1.0 & 4.1$\pm$1.4 & 3.2$\pm$1. & \textbf{2.4}$\pm$1.1 & 2.8$\pm$0.4\\
time (s) $\downarrow$ & 78 & 171 & 374 & 1200 & 90 & 9 & 18 & 36 & 72\\

\bottomrule
\end{tabular}
\label{tab:errorcomp1}
\end{table*}

Tab.~\ref{tab:errorcomp1} displays the accuracy of these three methods across all three scenes over a regularly sampled volume of locations on and above the table (size: $75 \times 60 \times 22.5$~cm, with $300 \cdot 250 \cdot 90 = 6.75$M points).  ``Collision'' refers to the percentage of locations which are actually inside (GT $<$ 0) but predicted to be outside (SDF $>$ 0), thereby risking collision.
``Misclassification'' refers to the percentage of all points that are misclassified as either inside or outside when they are actually outside or inside, respectively. ``Mean and max L1 error'' refers to the averaged or maximal L1 error between GT and method's output in the collision area.
Note that even our slowest method (8k iterations) is faster than the fastest competitor, and our worst method (1k iterations) achieves results that are better than almost all competitors.

Fig.~\ref{fig:accplot} compares the accuracy of the SDF computed by our method to the baselines.  Note that the proposed method is significantly more accurate.

\subsection{Manipulator Control}

Finally, we test the full integrated system including perception and control. This experiment is an important step toward validating our approach's ability of content creation (that is, reconstructing a mesh of the scene from sensory input), as well as its ability to perform real-to-sim transfer. 

\begin{table}
\centering
\caption{Using the estimated SDF for manipulator control.}
\begin{tabular}{l rrrr}
\toprule
Metric & Ours & COLMAP & iSDF & Voxblox \\
\cmidrule(r){2-5}
max penetration (cm) $\downarrow$ & \textbf{0.0} & \textbf{0.0} & 0.8 & 0.5\\
goal reaching error (cm) $\downarrow$ & \textbf{0.3} & 5.3 & 1.0 & 1.5\\
success rate ($\%$) $\uparrow$ & \textbf{85} & 35 & 5 & 5\\
\bottomrule
\end{tabular}
\label{tab:errorcomp2}
\end{table}

We randomly select 20 start and goal pairs at a fixed height above the table of Scene~1 and check whether the model predictive controller can generate a trajectory to attain that pose using the ground truth mesh for collision checking. We then swap out the ground truth mesh collision checker with the methods discussed in the previous section and run the robot through the same set of poses. We use a buffer of 1~cm for all methods (i.e., during planning the robot is maintained at a distance of at least 1~cm from the objects). Between each start and goal pose, the robot arm executes 500 timesteps at a rate of 50~Hz. Tab.~\ref{tab:errorcomp2} summarizes quantitative evaluation of the various methods. 
``Max penetration'' means the maximal distance that the robot arm collides into objects, ``goal reaching error'' refers to the Euclidean distance between the goal pose and the robot's final pose at the end of execution, and ``success rate'' refers to the percentage of runs for which the robot moves to within 1~cm of the goal pose without collision. Our method achieves higher accuracy, lower error, and higher success rate than the baselines. Fig.~\ref{fig:gripper} compares the ability of our method to achieve a target pose in clutter, to the result obtained with the COLMAP baseline.  Because COLMAP's reconstruction is conservative (larger than ground truth), the model predictive controller is unable to reach the target pose, even though it properly avoids collision.

\section{CONCLUSION}

We have presented a system for control of a robot manipulator based on 3D reconstruction of a tabletop scene observed by an RGB camera from multiple views.
Our method utilizes a fast NeRF approach to infer a high-fidelity reconstruction, from which a mesh is inferred.
While we initially were concerned that the mesh might not be watertight, our experiments show that instant NGP~\cite{mueller2022siggraph:instantngp} is able to yield watertight meshes without postprocessing.  
From the output mesh, a Euclidean full SDF (ESDF) is constructed, which specifies for any point in space its signed distance to the nearest surface.
We show that the resulting ESDF is more accurate than competing baselines, and we also demonstrate that the ESDF can be used to accurately control a robot manipulator to reach target poses in clutter without collision.
Future work should be aimed at incorporating motion generation and interaction with the scene.

\bibliographystyle{IEEEtran}
\bibliography{main}

\begin{thebibliography}{10}
\providecommand{\url}[1]{#1}
\csname url@samestyle\endcsname
\providecommand{\newblock}{\relax}
\providecommand{\bibinfo}[2]{#2}
\providecommand{\BIBentrySTDinterwordspacing}{\spaceskip=0pt\relax}
\providecommand{\BIBentryALTinterwordstretchfactor}{4}
\providecommand{\BIBentryALTinterwordspacing}{\spaceskip=\fontdimen2\font plus
\BIBentryALTinterwordstretchfactor\fontdimen3\font minus
  \fontdimen4\font\relax}
\providecommand{\BIBforeignlanguage}[2]{{%
\expandafter\ifx\csname l@#1\endcsname\relax
\typeout{** WARNING: IEEEtran.bst: No hyphenation pattern has been}%
\typeout{** loaded for the language `#1'. Using the pattern for}%
\typeout{** the default language instead.}%
\else
\language=\csname l@#1\endcsname
\fi
#2}}
\providecommand{\BIBdecl}{\relax}
\BIBdecl

\bibitem{schulman2014motion}
J.~Schulman, Y.~Duan, J.~Ho, A.~Lee, I.~Awwal, H.~Bradlow, J.~Pan, S.~Patil,
  K.~Goldberg, and P.~Abbeel, ``Motion planning with sequential convex
  optimization and convex collision checking,'' \emph{International Journal of
  Robotics Research}, 2014.

\bibitem{morgan2021vision}
A.~S. Morgan, B.~Wen, J.~Liang, A.~Boularias, A.~M. Dollar, and K.~Bekris,
  ``Vision-driven compliant manipulation for reliable, high-precision assembly
  tasks,'' \emph{RSS}, 2021.

\bibitem{mainprice2016warping}
J.~Mainprice, N.~Ratliff, and S.~Schaal, ``Warping the workspace geometry with
  electric potentials for motion optimization of manipulation tasks,'' in
  \emph{IROS}, 2016.

\bibitem{muglikar2020voxel}
M.~Muglikar, Z.~Zhang, and D.~Scaramuzza, ``Voxel map for visual {SLAM},'' in
  \emph{ICRA}, 2020.

\bibitem{mitash2020task}
C.~Mitash, R.~Shome, B.~Wen, A.~Boularias, and K.~Bekris, ``Task-driven
  perception and manipulation for constrained placement of unknown objects,''
  \emph{IEEE Robotics and Automation Letters}, vol.~5, no.~4, pp. 5605--5612,
  2020.

\bibitem{takikawa2021neural}
T.~Takikawa, J.~Litalien, K.~Yin, K.~Kreis, C.~Loop, D.~Nowrouzezahrai,
  A.~Jacobson, M.~McGuire, and S.~Fidler, ``Neural geometric level of detail:
  Real-time rendering with implicit 3d shapes,'' in \emph{CVPR}, 2021.

\bibitem{yu2021plenoctrees}
A.~Yu, R.~Li, M.~Tancik, H.~Li, R.~Ng, and A.~Kanazawa, ``Plenoctrees for
  real-time rendering of neural radiance fields,'' in \emph{ICCV}, 2021.

\bibitem{niessner2013real}
M.~Nie{\ss}ner, M.~Zollh{\"o}fer, S.~Izadi, and M.~Stamminger, ``Real-time 3d
  reconstruction at scale using voxel hashing,'' \emph{ACM Transactions on
  Graphics}, 2013.

\bibitem{mueller2022siggraph:instantngp}
T.~M\"uller, A.~Evans, C.~Schied, and A.~Keller, ``Instant neural graphics
  primitives with a multiresolution hash encoding,'' \emph{SIGGRAPH}, 2022.

\bibitem{sterzentsenko2019self}
V.~Sterzentsenko, L.~Saroglou, A.~Chatzitofis, S.~Thermos, N.~Zioulis,
  A.~Doumanoglou, D.~Zarpalas, and P.~Daras, ``Self-supervised deep depth
  denoising,'' in \emph{CVPR}, 2019.

\bibitem{chaudhary2016noise}
R.~Chaudhary and H.~Dasgupta, ``An approach for noise removal on depth
  images,'' in \emph{arXiv:1602.05168}, 2016.

\bibitem{sweeney2019icra}
C.~Sweeney, G.~Izatt, and R.~Tedrake, ``A supervised approach to predicting
  noise in depth images,'' in \emph{ICRA}, 2019.

\bibitem{mallick2014sensors}
T.~Mallick, P.~Das, and A.~Majumdar, ``Characterizations of noise in {K}inect
  depth images: A review,'' \emph{Sensors}, 2014.

\bibitem{haider2022sensors}
A.~Haider and H.~Hel-Or, ``What can we learn from depth camera sensor noise?''
  \emph{Sensors}, 2022.

\bibitem{handa:etal:2014}
A.~Handa, T.~Whelan, J.~McDonald, and A.~J. Davison, ``A benchmark for {RGB-D}
  visual odometry, {3D} reconstruction and {SLAM},'' in \emph{ICRA}, 2014.

\bibitem{baruhov2020gan}
A.~Baruhov and G.~Gilboa, ``Unsupervised enhancement of real-world depth images
  using tri-cycle {GAN},'' in \emph{arXiv:2001.03779}, 2020.

\bibitem{nguyen2012kinect}
C.~V. Nguyen, S.~Izadi, and D.~Lovell, ``Modeling {K}inect sensor noise for
  improved {3D} reconstruction and tracking,'' in \emph{International
  Conference on 3D Imaging, Modeling, Processing, Visualization \&
  Transmission}, 2012.

\bibitem{iversen2017kinect}
T.~Iversen and D.~Kraft, ``Generation of synthetic {K}inect depth images based
  on empirical noise model,'' \emph{Electronics Letters}, 2017.

\bibitem{sajjan2020clear}
S.~Sajjan, M.~Moore, M.~Pan, G.~Nagaraja, J.~Lee, A.~Zeng, and S.~Song, ``Clear
  grasp: 3d shape estimation of transparent objects for manipulation,'' in
  \emph{ICRA}, 2020.

\bibitem{gu2017learning}
S.~Gu, W.~Zuo, S.~Guo, Y.~Chen, C.~Chen, and L.~Zhang, ``Learning dynamic
  guidance for depth image enhancement,'' in \emph{CVPR}, 2017.

\bibitem{lin2021iros:mvml}
Y.~Lin, J.~Tremblay, S.~Tyree, P.~A. Vela, and S.~Birchfield, ``Multi-view
  fusion for multi-level robotic scene understanding,'' in \emph{IROS}, 2021.

\bibitem{schoenberger2016cvpr:sfm}
J.~L. Sch\"{o}nberger and J.-M. Frahm, ``Structure-from-motion revisited,'' in
  \emph{CVPR}, 2016.

\bibitem{schoenberger2016eccv:mvs}
J.~L. Sch\"{o}nberger, E.~Zheng, M.~Pollefeys, and J.-M. Frahm, ``Pixelwise
  view selection for unstructured multi-view stereo,'' in \emph{ECCV}, 2016.

\bibitem{lorensen1987siggraph:marchcube}
W.~E. Lorensen and H.~E. Cline, ``Marching cubes: A high resolution {3D}
  surface construction algorithm,'' in \emph{SIGGRAPH}, 1987.

\bibitem{oleynikova2017iros:voxblox}
H.~Oleynikova, Z.~Taylor, M.~Fehr, J.~Nieto, and R.~Siegwart, ``Voxblox:
  Incremental {3D} euclidean signed distance fields for on-board {MAV}
  planning,'' in \emph{IROS}, 2017.

\bibitem{mildenhall2020eccv:nerf}
B.~Mildenhall, P.~P. Srinivasan, M.~Tancik, J.~T. Barron, R.~Ramamoorthi, and
  R.~Ng, ``{NeRF}: Representing scenes as neural radiance fields for view
  synthesis,'' in \emph{ECCV}, 2020.

\bibitem{bhardwaj2021corl:storm}
M.~Bhardwaj, B.~Sundaralingam, A.~Mousavian, N.~D. Ratliff, D.~Fox, F.~Ramos,
  and B.~Boots, ``Storm: An integrated framework for fast joint-space
  model-predictive control for reactive manipulation,'' in \emph{CoRL}, 2022.

\bibitem{bloesch2018cvpr:codeslam}
M.~Bloesch, J.~Czarnowski, R.~Clark, S.~Leutenegger, and A.~J. Davison,
  ``{CodeSLAM} - learning a compact, optimisable representation for dense
  visual {SLAM},'' in \emph{CVPR}, 2018.

\bibitem{wada2020cvpr:morefusion}
K.~Wada, E.~Sucar, S.~James, D.~Lenton, and A.~J. Davison, ``Morefusion:
  Multi-object reasoning for 6d pose estimation from volumetric fusion,'' in
  \emph{CVPR}, 2020.

\bibitem{zimny2022arx:pointstonerf}
D.~Zimny, T.~Trzci{\'n}ski, and P.~Spurek, ``{Points2NeRF}: Generating neural
  radiance fields from {3D} point cloud,'' in \emph{arXiv:2206.01290}, 2022.

\bibitem{xu2022cvpr:pointnerf}
Q.~Xu, Z.~Xu, J.~Philip, S.~Bi, Z.~Shu, K.~Sunkavalli, and U.~Neumann,
  ``{Point-NeRF}: Point-based neural radiance fields,'' in \emph{CVPR}, 2022.

\bibitem{mescheder2019cvpr:occnet}
L.~Mescheder, M.~Oechsle, M.~Niemeyer, S.~Nowozin, and A.~Geiger, ``Occupancy
  networks: Learning {3D} reconstruction in function space,'' in \emph{CVPR},
  2019.

\bibitem{ortiz2022rss:isdf}
J.~Ortiz, A.~Clegg, J.~Dong, E.~Sucar, D.~Novotny, M.~Zollhoefer, and
  M.~Mukadam, ``isdf: Real-time neural signed distance fields for robot
  perception,'' in \emph{RSS}, 2022.

\bibitem{yariv2021neurips:volsdf}
L.~Yariv, J.~Gu, Y.~Kasten, and Y.~Lipman, ``Volume rendering of neural
  implicit surfaces,'' in \emph{NeurIPS}, 2021.

\bibitem{wang2021neurips:neus}
P.~Wang, L.~Liu, Y.~Liu, C.~Theobalt, T.~Komura, and W.~Wang, ``{NeuS}:
  Learning neural implicit surfaces by volume rendering for multi-view
  reconstruction,'' in \emph{NeurIPS}, 2021.

\bibitem{azinovic2022cvpr:nrgbd}
D.~Azinovi{\'c}, R.~Martin-Brualla, D.~B. Goldman, M.~Nie{\ss}ner, and
  J.~Thies, ``Neural {RGB-D} surface reconstruction,'' in \emph{CVPR}, 2022.

\bibitem{takikawa2021cvpr:nglod}
T.~Takikawa, J.~Litalien, K.~Yin, K.~Kreis, C.~Loop, D.~Nowrouzezahrai,
  A.~Jacobson, M.~McGuire, and S.~Fidler, ``Neural geometric level of detail:
  Real-time rendering with implicit {3D} shapes,'' in \emph{CVPR}, 2021.

\bibitem{yu2022arx:monosdf}
Z.~Yu, S.~Peng, M.~Niemeyer, T.~Sattler, and A.~Geiger, ``{MonoSDF}: Exploring
  monocular geometric cues for neural implicit surface reconstruction,'' in
  \emph{arXiv:2206.00665}, 2022.

\bibitem{sitzmann2020neurips:metasdf}
V.~Sitzmann, E.~R. Chan, R.~Tucker, N.~Snavely, and G.~Wetzstein, ``{MetaSDF}:
  Meta-learning signed distance functions,'' in \emph{NeurIPS}, 2020.

\bibitem{park2019cvpr:deepsdf}
J.~J. Park, P.~Florence, J.~Straub, R.~Newcombe, and S.~Lovegrove, ``{DeepSDF}:
  Learning continuous signed distance functions for shape representation,'' in
  \emph{CVPR}, 2019.

\bibitem{ratliff2018arx:rmp}
N.~D. Ratliff, J.~Issac, D.~Kappler, S.~Birchfield, and D.~Fox, ``Riemannian
  motion policies,'' in \emph{arXiv:1801.02854}, 2018.

\bibitem{battaje2022iros:ooaat}
A.~Battaje and O.~Brock, ``One object at a time: Accurate and robust structure
  from motion for robots,'' in \emph{IROS}, 2022.

\bibitem{yang2022icra:mpc}
W.~Yang, B.~Sundaralingam, C.~Paxton, I.~Akinola, Y.-W. Chao, M.~Cakmak, and
  D.~Fox, ``Model predictive control for fluid human-to-robot handovers,'' in
  \emph{ICRA}, 2022.

\bibitem{adamkiewicz2022icra:vorn}
M.~Adamkiewicz, T.~Chen, A.~Caccavale, R.~Gardner, P.~Culbertson, J.~Bohg, and
  M.~Schwager, ``Vision-only robot navigation in a neural radiance world,'' in
  \emph{ICRA}, 2022.

\bibitem{driess2021corl:lmaf}
D.~Driess, J.-S. Ha, M.~Toussaint, and R.~Tedrake, ``Learning models as
  functionals of signed-distance fields for manipulation planning,'' in
  \emph{CoRL}, 2021.

\bibitem{zhi2022trob:diffco}
Y.~Zhi, N.~Das, and M.~Yip, ``{DiffCo}: Auto-differentiable proxy collision
  detection with multi-class labels for safety-aware trajectory optimization,''
  \emph{IEEE Transactions on Robotics}, 2022.

\bibitem{danielczuk2021icra:collfun}
M.~Danielczuk, A.~Mousavian, C.~Eppner, and D.~Fox, ``Object rearrangement
  using learned implicit collision functions,'' in \emph{ICRA}, 2021.

\bibitem{li2021corl:visuomotor}
Y.~Li, S.~Li, V.~Sitzmann, P.~Agrawal, and A.~Torralba, ``{3D} neural scene
  representations for visuomotor control,'' in \emph{CoRL}, 2021.

\bibitem{driess2022reinforcement}
D.~Driess, I.~Schubert, P.~Florence, Y.~Li, and M.~Toussaint, ``Reinforcement
  learning with neural radiance fields,'' \emph{arXiv preprint
  arXiv:2206.01634}, 2022.

\bibitem{pantic2022icraw:sfog}
M.~Pantic, C.~Cadena, R.~Siegwart, and L.~Ott, ``Sampling-free obstacle
  gradients and reactive planning in neural radiance fields,'' in \emph{ICRA
  Workshop on Motion Planning with Implicit Neural Representations of
  Geometry}, 2022.

\bibitem{mikhailov2019:turbo}
A.~Mikhailov, ``Turbo, an improved rainbow colormap for visualization,''
  \url{https://ai.googleblog.com/2019/08/turbo-improved-rainbow-colormap-for.html},
  2019.

\bibitem{wen2022you}
B.~Wen, W.~Lian, K.~Bekris, and S.~Schaal, ``You only demonstrate once:
  Category-level manipulation from single visual demonstration,'' \emph{RSS},
  2022.

\bibitem{izadi2011kinectfusion}
S.~Izadi, D.~Kim, O.~Hilliges, D.~Molyneaux, R.~Newcombe, P.~Kohli, J.~Shotton,
  S.~Hodges, D.~Freeman, A.~Davison \emph{et~al.}, ``{KinectFusion}: real-time
  {3D} reconstruction and interaction using a moving depth camera,'' in
  \emph{ACM Symposium on User Interface Software and Technology}, 2011.

\bibitem{hornung2013octomap}
A.~Hornung, K.~M. Wurm, M.~Bennewitz, C.~Stachniss, and W.~Burgard,
  ``{OctoMap}: An efficient probabilistic {3D} mapping framework based on
  octrees,'' \emph{Autonomous Robots}, 2013.

\bibitem{wen2022catgrasp}
B.~Wen, W.~Lian, K.~Bekris, and S.~Schaal, ``{CaTGrasp}: Learning
  category-level task-relevant grasping in clutter from simulation,'' in
  \emph{ICRA}, 2022.

\bibitem{schoenberger2016sfm}
J.~L. Sch\"{o}nberger and J.-M. Frahm, ``Structure-from-motion revisited,'' in
  \emph{CVPR}, 2016.

\bibitem{balakumar2023curobo}
B.~Sundaralingam, S.~Hari, A.~Fishman, C.~Garrett, K.~V. Wyk, V.~Blukis,
  A.~Millane, H.~Oleynikova, A.~Handa, F.~Ramos, N.~Ratliff, and D.~Fox,
  ``Curobo: Parellelized collision-free robot motion generation,'' in
  \emph{ICRA}, 2023.

\bibitem{tyree2022iros:hope}
S.~Tyree, J.~Tremblay, T.~To, J.~Cheng, T.~Mosier, J.~Smith, and S.~Birchfield,
  ``6-{DoF} pose estimation of household objects for robotic manipulation: An
  accessible dataset and benchmark,'' in \emph{IROS}, 2022.

\bibitem{labbe2020eccv:cosypose}
Y.~{Labbe}, J.~{Carpentier}, M.~{Aubry}, and J.~{Sivic}, ``{CosyPose}:
  Consistent multi-view multi-object {6D} pose estimation,'' in \emph{ECCV},
  2020.

\end{thebibliography}

\end{document}